\documentclass[sigconf, 9pt]{acmart}

\AtBeginDocument{%
  \providecommand\BibTeX{{%
    \normalfont B\kern-0.5em{\scshape i\kern-0.25em b}\kern-0.8em\TeX}}}

\setcopyright{acmcopyright}
\copyrightyear{2020}
\acmYear{2020}
\acmDOI{10.1145/1122445.1122456}

\acmConference[SenSys'20]{The 18th ACM Conference on Embedded Networked Sensor Systems}{November 16-19, 2020}{Virtual Event, Japan}
\acmBooktitle{Proceedings of The 18th ACM International Conference on Embedded
Networked Sensor Systems (SenSys), Nov 16-19, 2020, Virtual Event, Japan}
\acmPrice{15.00}
\acmISBN{978-1-4503-9999-9/18/06}

\usepackage{amsmath}
\DeclareMathOperator*{\argmax}{argmax}
\DeclareMathOperator*{\argmin}{argmin}

\usepackage{xspace}
\usepackage{multirow}
\usepackage{hyperref}
\usepackage[shortlabels]{enumitem}
\setlist[enumerate]{nosep}
\setlist{nolistsep,leftmargin=2.0mm}

\newcommand{\name}{{ApproxDet}\xspace} 
\newcommand{\eg}{{e.g.,}\xspace}
\newcommand{\ie}{{i.e.,}\xspace}
\newcommand{\et}{\textit{et al.}\xspace}
\newcommand{\iconsupport}{\includegraphics[width=0.17in]{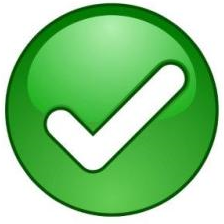}}
\newcommand{\iconpartsupport}{\includegraphics[width=0.17in]{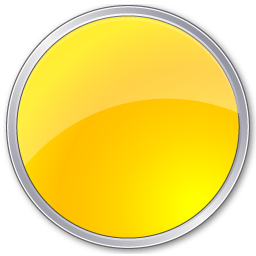}}
\newcommand{\iconnotsupport}{\includegraphics[width=0.17in]{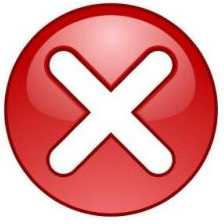}}
\newcommand{\chop}[1]{} 

\begin{document}
\title{\name: Content and Contention-Aware Approximate Object Detection for Mobiles}

\author{Ran Xu}
\affiliation{%
  \institution{Purdue University}
  }
\email{xu943@purdue.edu}

\author{Chen-lin Zhang}
\affiliation{%
  \institution{Nanjing University}
  }
\email{zhangcl@lamda.nju.edu.cn}

\author{Pengcheng Wang}
\affiliation{%
  \institution{Purdue University}
  }
\email{wang4495@purdue.edu}

\author{Jayoung Lee}
\affiliation{%
  \institution{Purdue University}
  }
\email{lee3716@purdue.edu}

\author{Subrata Mitra}
\affiliation{%
  \institution{Adobe Research}
  }
\email{subrata.mitra@adobe.com}

\author{Somali Chaterji}
\affiliation{%
  \institution{Purdue University}
  }
\email{schaterji@purdue.edu}

\author{Yin Li}
\affiliation{%
  \institution{University of Wisconsion-Madison}
  }
\email{yin.li@wisc.edu}

\author{Saurabh Bagchi}
\affiliation{%
  \institution{Purdue University}
  }
\email{sbagchi@purdue.edu}

\renewcommand{\shortauthors}{Xu et al.}

\begin{abstract}
Advanced video analytic systems, including scene classification and object detection, have seen widespread success in various domains such as smart cities and autonomous transportation. With an ever-growing number of powerful client devices, there is incentive to move these heavy video analytics workloads from the cloud to mobile devices to achieve low latency and real-time processing and to preserve user privacy.
However, most video analytic systems are heavyweight and are trained offline with some {\em pre-defined} latency or accuracy requirements. This makes them unable to adapt at runtime in the face of three types of dynamism --- the input video characteristics change, the amount of compute resources available on the node changes due to co-located applications, and the user's latency-accuracy requirements change. 
In this paper we introduce \name, an {\em adaptive} video object detection framework for mobile devices to meet accuracy-latency requirements in the face of changing content and resource contention scenarios. 
To achieve this, we introduce a multi-branch object detection kernel (layered on Faster R-CNN), which incorporates a data-driven modeling approach on the performance metrics, and a latency SLA-driven scheduler to pick the best execution branch at runtime. We couple this kernel with approximable video object tracking algorithms to create an end-to-end video object detection system.
We evaluate \name on a large benchmark video dataset and compare quantitatively to AdaScale and YOLOv3. We find that \name is able to adapt to a wide variety of contention and content characteristics and achieves 52\% lower latency and 11.1\% higher accuracy over YOLOv3, outshining all baselines.
\end{abstract}



\maketitle
\vspace{-0.5em}
\section{Introduction}
\label{sec_introduction}

Mobile devices with integrated cameras have seen tremendous success in various domains. Equipped with increasingly powerful System-on-Chips (SoCs), mobile augmented reality (AR) devices such as the Microsoft Hololens
and Magic Leap One,
along with top-of-the-line smartphones like iPhone 11 Pro and Samsung Galaxy S20, are opening up a plethora of new continuous mobile vision applications that were previously deemed impossible. These applications range from detection of objects around the environment for immersive experience in AR games such as Pokemon-Go~\cite{pokemongo}, to recognition of road signs for providing directions in real-time~\cite{chen2015glimpse}, to identification of people for interactive photo editing~\cite{shen2016automatic}, and to Manchester City's AR-driven stadium tour.
A fundamental vision task that all of these applications must perform, is object detection on the live video stream that the camera is capturing. 
To maintain the immersive experience of the user (\eg for AR games) or to give usable output on time (\eg for road sign recognition), such tasks must be performed in near real-time with very low latency.

Computer vision and computer systems research working together has made significant progress in lightweight object detection applicable to mobile settings for {\em still images} in recent years~\cite{wang2018pelee,redmon2018yolov3, liu2016ssd,tan2019efficientdet}, thanks to development of efficient deep neural networks (DNNs)~\cite{howard2017mobilenets,zhang2018shufflenet,han2015deep,hubara2017quantized}. However, directly applying image-based object detectors to video streams does not work well~\cite{zhao2019object}, especially in a mobile setting. \textit{First}, applying a detector on all video frames introduces excessive computational cost and would often violate the latency requirements of our target continuous vision applications.
\textit{Second}, image-based object detectors are not cognizant of the significant temporal continuity that exists in successive video frames (\eg a static scene with a slowly moving object) and therefore cannot leverage them for fitting in the latency budget. 
To overcome these algorithmic challenges, the computer vision community has proposed some DNN models~\cite{kang2017t,zhu2017flow,feichtenhofer2017detect} for video object detection and tracking. More recently, several lightweight DNN models~\cite{liu2019edge,zhu2018towards} that are suitable for mobile devices were developed.

\begin{figure}[t]
    \begin{minipage}[t]{0.49\linewidth} 
        \centering
        \includegraphics[width=1\columnwidth]{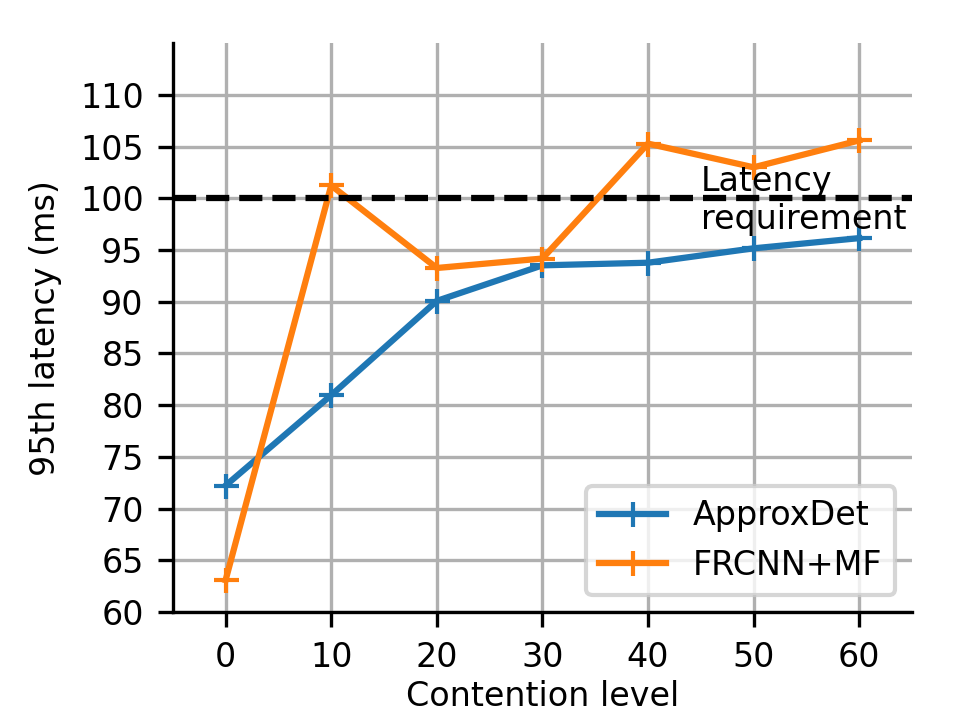}
    \end{minipage}
    \hfill
    \begin{minipage}[t]{0.49\linewidth} 
        \centering
        \includegraphics[width=1\columnwidth]{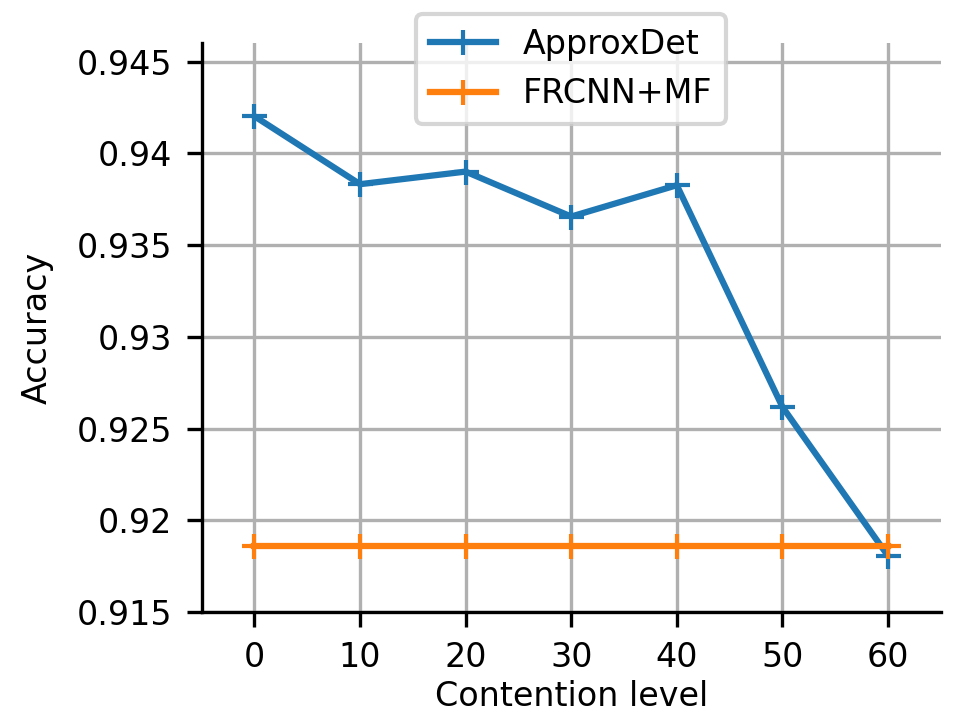}
    \end{minipage}
    \vspace{-1em}
    \caption{\textbf{\name}: The first system of mobile video object detection that takes both video content-awareness \textit{and} resource contention-awareness within its ambit. Compared to a widely used object detector~\cite{ren2015faster} optimized for a target latency requirement (orange curve), \name (blue curve) keeps its runtime latency (left) below the requirement and achieves better accuracy (right).}
    \label{fig:motivation_for contention}
    \vspace{-1.5em}
\end{figure}

Despite these efforts, we argue that the system challenges of video object detection for continuous vision applications on resource-constrained devices remain largely unsolved. 
A major shortcoming is that none of the existing approaches can adapt to runtime condition changes, such as {\em the content characteristics of the input videos}, and \textit{the level of contention on the edge device}. Modern mobile devices\footnote{In this paper, we use the term mobiles for the target platform, though without loss of generality, it applies to mobile \textit{and} embedded platforms. The commonality is that both are using increasingly powerful SoCs, but are still resource constrained relative to the servers where streaming video analytics are typically run. Further, both are used to run co-located applications (mobiles more so than embedded), which can interfere with the video analytics.} come with increasingly powerful SoCs having multiple heterogeneous processing units, and no longer process just a single application at a time. For example, both iOS and Android support multiple background tasks~\cite{background_ios, background_ios_react, background_android}, such as an always-on personal assistant like Siri running a DNN for speech recognition (GPU contention), or a firewall constantly inspecting packets (memory bandwidth contention). These tasks can run simultaneously with a continuous vision application that requires a video object detector, leading to unpredictable resource contention on mobile devices similar to a traditional server setting~\cite{xu2018pythia, delimitrou2013, maji2014, mars2011bubble}. Such concurrent applications or background tasks can compete for resources critical to object detection, thus drastically increasing the latency of the object detector. Consider the example of a widely used DNN-based object detector -- Faster R-CNN (FRCNN)~\cite{ren2015faster}, integrated with MedianFlow (MF) object tracking~\cite{MedianFlow} and optimized for a latency requirement of 100 milliseconds (ms)\footnote{We pick a combination of detector and tracker configurations that satisfies the latency requirement in 95\% of the video frames in a validation dataset.}. The orange curve in Figure~\ref{fig:motivation_for contention} shows the latency (left) and accuracy (right) of this FRCNN+MF processing an input video at different GPU contention levels on an embedded device (NVIDIA TX2). Without contention, the detector has a latency of $\approx$ 64 ms. However, as the GPU contention level increases, we observe a severe increase in the detection latency. While the accuracy remains the same, the latency of the detector fluctuates significantly and violates our latency requirement of 100 ms. Different from server-class devices~\cite{xu2018pythia}, our target mobile devices have less ability to isolate co-located applications from interference from one another. This happens due to the paucity of isolation mechanisms like VMs with resource reservation on our target class of mobile devices. 

To address this issue, we propose \textbf{\name}, a novel system that takes both video content-awareness \textit{and} resource contention-awareness within its ambit. 
In contrast to the static FRCNN+MF baseline in Figure~\ref{fig:motivation_for contention}, \name manages to keep a latency below the requirement with increased level of contention while achieving a better accuracy. 
To this end, \name uses a {\em single} model with multiple approximation knobs that are dynamically tuned at runtime to stay on the Pareto optimal frontier of the latency/accuracy tradeoff curve. We refer to the execution branch with a particular configuration of the approximation knob as an {\bf {\em approximation branch (AB)}}.

This overall functionality is supported by {\em three core technical contributions} of this work. {\em First}, \name models the impacts of the contention level to the latency of the ABs. {\em Second}, our model combines an offline trained latency prediction model and an online contention sensor to precisely predict the latency of each AB in our system. Thus \name can adapt to resource contention at a given latency budget at runtime, an ability especially critical for the deployment on edge devices as their resources are limited and shared.
{\em Third}, \name further considers how the video content influences both accuracy and latency. \name leverages video characteristics such as the object motion (fast vs. slow) and the sizes and the number of objects, to better predict the accuracy and latency of the ABs, and to select the best AB with reduced latency and increased accuracy. 

Figure~\ref{fig:workflow} presents an overview of \name. The object detection pipeline comprises of an object detector DNN based on our modified version of FRCNN~\cite{ren2015faster}, and a video object tracker that can be selected from among a set of choices. Both the detector and tracker are {\em jointly} approximated to achieve the required point in the accuracy-latency trade-off curve. 
Importantly, with the joint modeling of resource contention and content characteristics, \name can dynamically tune the approximation knobs, including the interval of performing object detection on video frames, the input shape of the frames and the number of proposals in the object detector DNN, the choice of object trackers, and the down-sampling ratio of the tracker.
This means, in response to a resource contention from the GPU, \name can move to a more aggressive approximation setting of the detector DNN to bring down the latency since the detector DNN is more sensitive to the GPU contention. 
Moreover, in case of a content change in the video frame, \eg a rapidly moving object, \name is able to switch over to a different AB where detector DNN is triggered more frequently to mitigate the tracker failure due to the fast-moving objects.

To our best knowledge, \name is the first system that accounts for \textit{the joint adaptation to video content and resource contention} for mobile vision applications. 
Distinct from prior work that optimizes multiple concurrent DNN applications~\cite{jiang2018mainstream, fang2018nestdnn, mcdnn}, our system treats the contention as a black box. Our principle is that we neither know the context nor have control over the contention in real-world systems, as these video-analytic systems are typically user-space processes without any OS privilege. \chop{Further, \name considers the general source of contention not limited to deep learning applications.} To estimate the contention, \name uses the current observed latency to map to the contention level, and adapts to use the AB that can satisfy the latency requirement from the user. Our evaluation bears out that this design choice is particularly effective for handling varied forms of contention under one simple algorithm.
Table~\ref{table:comparison} further contrasts the features of \name with existing works.

To evaluate our model, we conduct extensive experiments on ImageNet Video Object Detection (VID) dataset and compare our \name to a number of baselines, including AdaScale~\cite{chin2019adascale}, Faster R-CNN~\cite{ren2015faster}, Faster R-CNN with tracking~\cite{MedianFlow}, and YOLOv3~\cite{redmon2018yolov3}. Our results suggest that \name is able to adapt to a wide variety of contention and content characteristics and achieves 52\% lower latency and 11.1\% higher accuracy over the latest YOLOv3 optimized for efficiency and accuracy, and outshining all other baselines.

\begin{table}[t]
  \centering
  \caption{A comparison of the key features of our \name solution to previous approaches. \name provides the most flexible framework for adaptive video object detection. ``Single/multi model'': if a method uses shared execution branch for different control parameters. ``Switching cost considered'': a technique takes into account switching cost while making its decision.}
  \label{table:comparison}
  \vspace{-0.5em}
  \scalebox{0.7}{
  \begin{tabular}{|p{0.63in}|  p{0.35in}| p{0.5in}|p{0.5in}|p{0.5in} | p{0.3in}|p{0.45in}|p{0.3in}|}
    \hline
    Solution & Single/ Multi Model (S/M)           & Tuning Knobs  
             & Switching cost considered           & Dynamic Scenario   
             & Open Source                    & Mobile/ Server (M/S) 
             & Video/ Image                \\
    \hline
    \name    & S              
             & si, shape, nprop, tracker, and ds
             & \iconsupport            & \iconsupport
             & \iconsupport            & M              
             & VID                     \\  
    \hline
    Faster R-CNN & S
            & shape, nprop
            & \iconnotsupport & \iconnotsupport 
            & \iconsupport & S
            & VID \\
    \hline
    YOLO, SSD & S
            & shape
            & \iconnotsupport & \iconnotsupport
            & \iconsupport & S
            & VID \\
    \hline    
    AdaScale
             & S & scale 
             & \iconnotsupport & \iconpartsupport 
             & \iconsupport & S  
             & VID                     \\ 
    \hline
    MCDNN [MobiSys16] 
             & M & models& \iconnotsupport 
             & \iconnotsupport  & \iconnotsupport & M+S
             & IMG \\ 
    \hline
    NestDNN [MobiCom18] 
             & S & \# filters &  \iconpartsupport 
             & \iconnotsupport & \iconnotsupport 
             & M & IMG \\ 
    \hline
    RANet [CVPR20]
             & M             & resource budget
             & \iconnotsupport         & \iconnotsupport
             & \iconsupport            & S                   
             & IMG        \\             
    \hline
    \multicolumn{8}{|c|}{\iconsupport Supported \iconpartsupport Partially Supported \iconnotsupport~~Not Supported} \\ \hline
  \end{tabular}
  }
  \vspace{-1em}
\end{table}

In summary, our work makes the following contributions:
\begin{enumerate}
     \item We show that contention in mobile/embedded devices can significantly degrade the latency requirements of continuous vision applications.
     \item We propose \name --- an adaptive object detection framework that takes the runtime content characteristics \textit{and} resource availability into consideration to dynamically optimize for the best accuracy-vs-latency tradeoff. This optimization is done using a single model with different approximation branches, rather than using an ensemble of models, reducing switching overhead and the memory footprint.
     \item \name makes use of video-specific features, \ie it does not consider video to be simply a set of discrete image frames. For example, it uses near past video content characteristics (such as, size of the objects) to guide its choice of the optimal approximation branch. 
\end{enumerate}
\vspace{-0.5em}
\section{Background}
\label{sec_background}

This section introduces the background of \name, including object detection and tracking models used in \name and the context of approximate computing on edge devices.

\vspace{-0.5em}
\subsection{Object Detection}

Given an input image or video frame, an object detector aims at locating tight bounding boxes of object instances from target categories. 
In terms of network architecture, a CNN-based object detector can be divided into the backbone part that extracts image features, and the detection part that classifies object regions based on the extracted features. The detection part can be further divided into two-stage~\cite{ren2015faster,dai2016r} and single-stage detectors~\cite{liu2016ssd,redmon2018yolov3,lin2017focal}. Two-stage detectors usually make use of Region Proposal Networks (RPN) for generating regions-of-interest (RoIs), which are further refined through the detection head and thus more accurate.

Our work builds on Faster-RCNN~\cite{ren2015faster} --- an accurate and flexible framework for object detection and a canonical example of a two-stage object detector. 
An input image or video frame is first resized to a specific \emph{input shape} and fed into a DNN, where image features are extracted. Based on the features, a RPN identifies a \emph{pre-defined number} of candidate object regions, known as region proposals. Image features are further aggregated within the proposed regions, followed by another DNN to classify the proposals into either background or one of target object categories and to refine the location of the proposals. Our key observation is that the input shape and the number of proposals have significant impact to the accuracy and latency. Therefore, we propose to expose \emph{input shape} and \emph{number of region proposals} as tuning knobs in \name.

Another line of research develops single-stage object detection~\cite{liu2016ssd, redmon2018yolov3}. Without using region proposals, these models are optimized for efficiency and oftentimes less flexible. We consider one of the representative single-stage detectors as a baseline~\cite{redmon2018yolov3} in our experiments. YOLO simplifies object detection as a regression problem by directly predicting bounding boxes and class probabilities without the generation of region proposals. 

\begin{figure*}[t]
    \centering
    \includegraphics[width=0.85\linewidth]{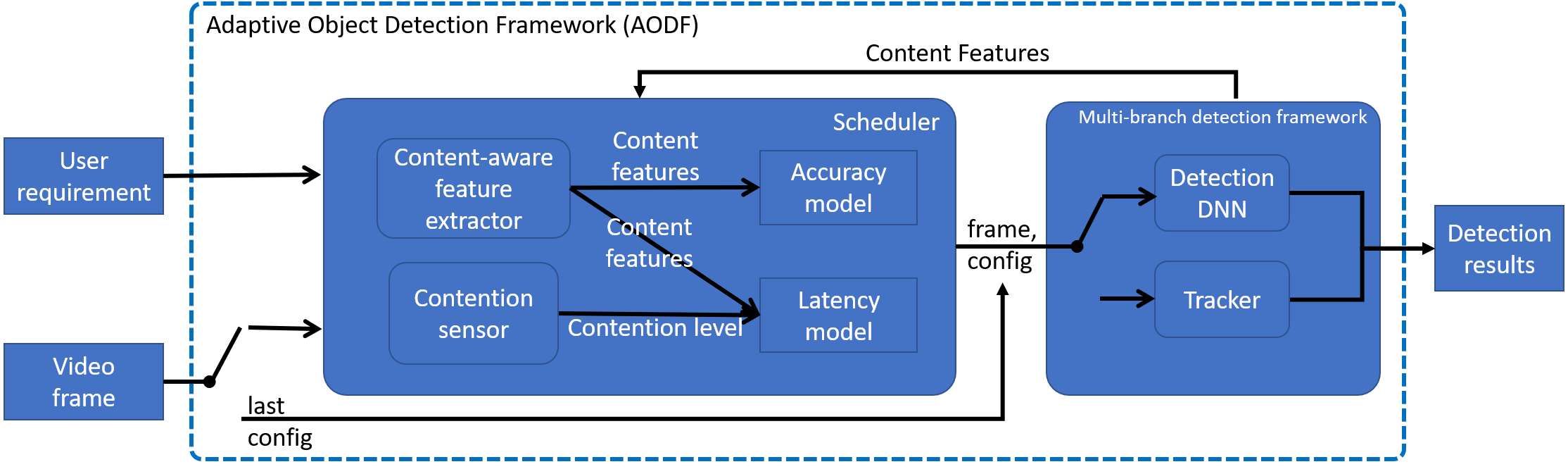}
    \vspace{-0.5em}
    \caption{The workflow of the adaptive object detection framework.}
    \label{fig:workflow}
    \vspace{-1em}
\end{figure*}

\subsection{Object Tracking}
Object tracking seeks to locate moving objects over time within a video. We focused on motion-based visual tracking due to its simplicity and efficiency. A motion-based tracker assumes the initial position of each object is given in a starting frame, and makes use of local motion cues to predict the object's position in the next batch of frames. Our system considers a set of existing motion-based object trackers --- MedianFlow~\cite{MedianFlow}, KCF~\cite{KCF}, and CSRT~\cite{CSRT}. The key difference lies in the extraction of motion cues, via \eg optical flow or correlation filters, leading to varying accuracy and efficiency under different application scenarios. We thus propose to enable the adaptive \emph{choice of the trackers} as one of our tuning knobs.

Another important factor of object tracking performance is the input resolution to a motion-based tracker. A downsampled version of the input image allows to better capture large motion and thus to track fast-moving objects, while a high-resolution input image facilitates the accurate tracking of objects that move slowly. Therefore, we further expose \emph{the downsampling ratio of the input image} as another tuning knob for tracking.

\vspace{-0.5em}
\subsection{Approximate Computing and Adaptation}

Many computations are inherently approximate---they trade off quality of results for lower execution time or lower energy. Approximate computing has emerged as an area that exposes additional sources of approximation at the computer system level, including resource-constrained mobile and embedded platforms. One challenge in approximate computing is that the accuracy and performance of applying approximate techniques to a specific application and input sets are hard to predict and control~\cite{mitra2017phase, ding2015autotuning, ansel2011language}. This may lead to missed optimization opportunities, unacceptable quality outputs, and even incorrect executions. The two fundamental causes is that approximation techniques are not content-aware and contention-aware. Some recent work has started to address these issues. For example, Input Responsive Approximation (IRA)~\cite{laurenzano2016input} and VideoChef~\cite{xu2018videochef} have brought in content-aware approximation for image processing and video processing pipelines respectively. 

To the best of our knowledge there is no solution that makes video analytics on mobile platforms adaptive to resource contention. 
Recently, Min \textit{et al.}~\cite{min2019closer} assess the runtime quality of sensing models in the multi-mobile-device environment so that the best device is selected as a function of model accuracy. However, we have not seen similar work on the video object detection task.
There are several works that provide tunable knobs to trade off accuracy-versus-latency, primarily in the image processing context~\cite{huang2017multi,fang2018nestdnn,yang2020resolution}, and some in video processing context~\cite{chin2019adascale}. It is conceivable that these knobs can be reconfigured to an optimal setting continuously as contention varies. However, there are key systems challenges that have to be solved before that end goal can be achieved. Such challenges include how to sense contention, how to change the knob in response to a specific level of contention, and how to optimize for the switching overhead from one approximation level to another.
\vspace{-0.5em}
\section{Overview}
\label{sec_overview}

Figure~\ref{fig:workflow} presents the overall workflow of our system \name. 
Our system consists of two modules --- a scheduler and a multi-branch object detection framework. The detection framework takes a video frame and a configuration as an input and produces the detection results while the scheduler decides which configuration the detection framework should use. 

The detection framework includes two kernels: a detection kernel and a tracking kernel. This follows the common practice for video object detection that combines the heavy-weight detection and the light-weight tracker~\cite{liu2019edge, zhu2017deep}. At a high-level, the detection framework exposes five tuning knobs. With each tuning knob varying in a dynamic range, we construct a multi-dimensional configuration space and call the execution path of each configuration {\em {\bf an approximation branch (AB)}}
The accuracy and the latency (execution time) are different for each AB and the values depend upon the video content characteristics (\eg still versus fast-moving) and the compute resources available (\eg lightly-loaded versus heavily-loaded mobile). 
To efficiently select an AB at runtime according to the given (and possibly changing) user requirement, the scheduler estimates the current latency and accuracy of each branch. 
The scheduler then selects the most accurate/fastest branch according to the specific user requirement.

We train an accuracy model and a latency model offline to support such estimation online. 
To better predict such online performance metric, we build two lightweight online modules -- (1) a \textit{content-aware feature extractor}, which extracts the height, width, tracks the object information of the last frame, and calculates the object movements of the past few frames, and (2) a \textit{contention sensor}, which senses the current resource contention level. 
The scheduler is designed to run occasionally to re-calibrate the best approximation branch based on a learnable interval called ``scheduler interval'', which represents the number of frames that the configuration of the detection framework can be maintained. 
\section{Design Elements of \name}
\label{sec_technique}

\subsection{Multi-branch Object Detection Framework}
\label{subsec_framework}

To support the runtime adaptive object detection framework on videos, we first design a multi-branch object detection framework, with light switching overheads between different branches so that it can quickly adapt to runtime changes.
Different from object detection on still images, videos have temporal similarities and an object tracker is widely used to reduce the runtime cost with minor accuracy drop. Similarly, we compose an object detection DNN and an object tracker. The detection DNN produces initial bounding boxes for each object in the input image, while the tracker tracks objects between successive frames.

\begin{figure}[t]
    \centering
    \includegraphics[width=0.8\linewidth]{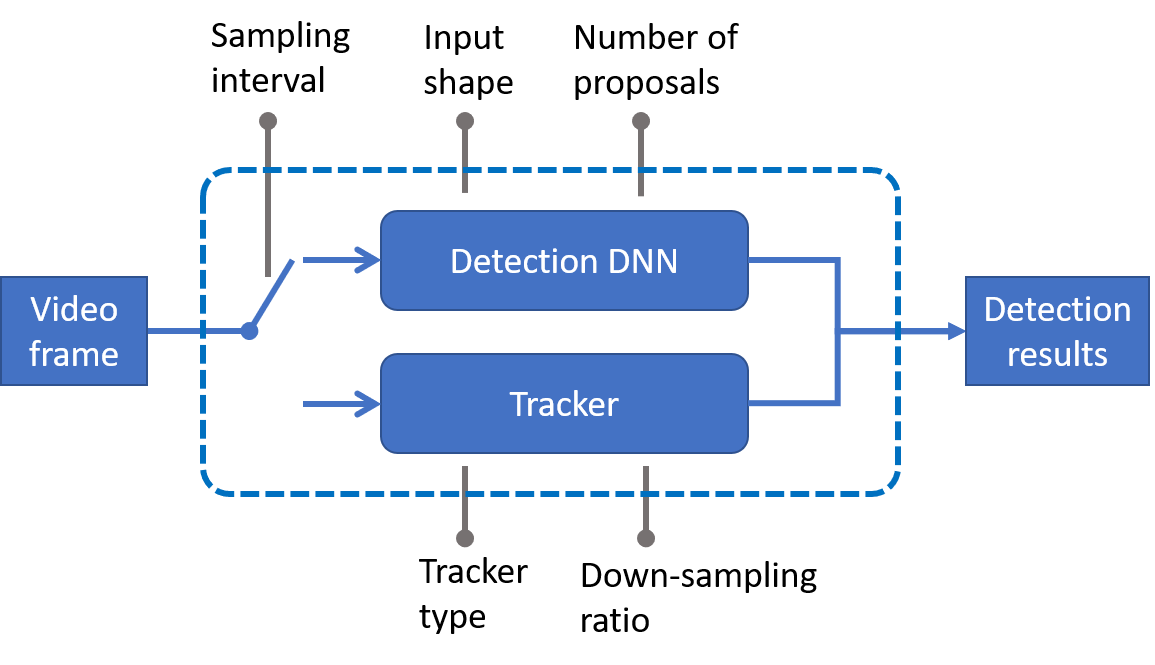}
    \vspace{-1em}
    \caption{Our multi-branch object detection framework \name with the five runtime tuning knobs.}
    \label{fig:detection_framework}
    \vspace{-1em}
\end{figure}

The overwhelming majority of work on lightweight object detection (such as that suited for our target class of devices, mobiles or embedded devices) is for images, such as YOLOv3~\cite{redmon2018yolov3} and SSD~\cite{liu2016ssd} and thus, does not leverage the video characteristics inherent in relation across image frames. For us, this leads to several insights and guides several design decisions. 

For the detection DNN, we choose the popular Faster-RCNN with ResNet-50 as the backbone~\cite{ren2015faster}. It shows state-of-the-art performance with medium speed when compared with other detection models. 
For the tracker part, we experiment with a set of 4 trackers --- MedianFlow~\cite{MedianFlow}, KCF~\cite{KCF}, CSRT~\cite{CSRT}, and Dense Optical Flow~\cite{farneback2003two}. These trackers are open-sourced in OpenCV with reasonable performance.
We then expose \textit{five tuning knobs} for this object detection framework that our scheduler controls programmatically at runtime to achieve the right accuracy-latency tradeoff. We introduce them below and illustrate them in Figure~\ref{fig:detection_framework}.

\begin{itemize}
    \item Sampling interval ($si$): For every $si$ frame, we run the heavyweight object detection DNN on the first frame and light-weight object tracker on the rest of the frames. 
    \item Input shape ($shape$): The resized shape 
    of the video frame that is fed into the detection DNN.
    \item Number of proposals ($nprop$): The number of proposals generated from the Region Proposal Networks (RPN) in our detection DNN.
    \item Tracker type ($tracker$): Type of object tracker.  
    \item Down-sampling ratio ($ds$): The downsampling ratio of the frame used by the object tracker.
\end{itemize}

Generally, we have empirically observed that smaller $si$, larger $shape$, more $nprop$, and smaller $ds$ will raise the accuracy and vice-versa. We will discuss the specifics of the knobs in Section~\ref{sec:implementation_values}.

\vspace{-0.5em}
\subsection{Content Feature Extraction}

As multi-branch object detection framework is designed, an important prerequisite is to {\em precisely estimate the accuracy and computation time (latency) of each approximation branch}.
To start with, the content feature has great impact on both the accuracy and latency of each AB based on the following two observations -- (1) tracker latency is affected by the number and area of the objects because tracker algorithms take the bounding boxes of the detection frames as inputs and calculate features inside each box; (2) both detection and tracker accuracy are affected by the content in the video. For example, detection DNNs perform consistently poorly with small objects on MS COCO dataset, including Faster-RCNN~\cite{ren2015faster}, SSD~\cite{liu2016ssd}, and YOLO~\cite{redmon2016you}. Moreover, both detection DNN and tracker find it harder to deal with fast-moving objects. Some previous works~\cite{chen2015glimpse} mention that movement between frames can be used as a feature to trigger the heavy detection process. This implies that for video object detection systems, we need to
extract these content features to improve the accuracy and latency of our models. In this paper, we mainly consider two types of content features, described next.

\subsubsection{Object Basic Features} 
\label{sec:object_feats}

We use {\em the number of objects} and {\em the summed area of the objects} as features for modeling the tracker latency. 
The intuition is that some light-weight trackers' latency increases proportionally with the number of objects and the area of the objects since each object is tracked independently, and the larger the area, the more tracking-related features need computation. These features can be easily extracted with no extra cost from the output of the detection DNN.

\begin{figure*}[h]
    \begin{minipage}[t]{0.3\linewidth} 
        \centering
        \includegraphics[width=1.05\linewidth]{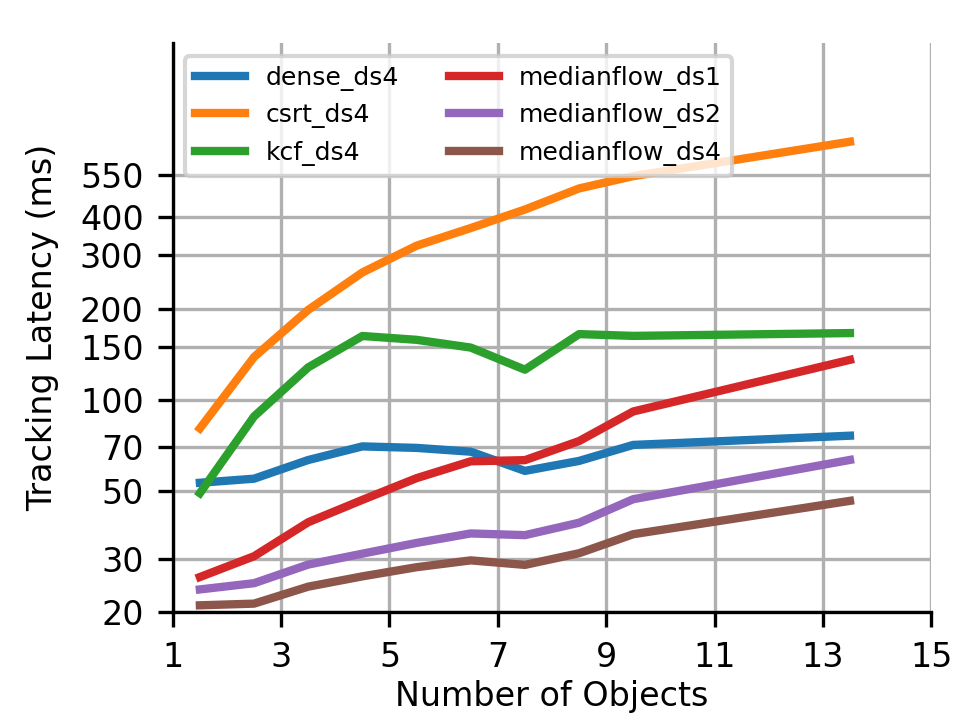}
        \vspace{-1em}
        \caption{The latency curve of object trackers on different numbers of the objects on the validation dataset.}
        \label{fig:tracker_latency_nobjs_feature}
        \vspace{-0.5em}
    \end{minipage}
    \hfill 
    \begin{minipage}[t]{0.3\linewidth} 
        \centering
        \includegraphics[width=1.05\linewidth]{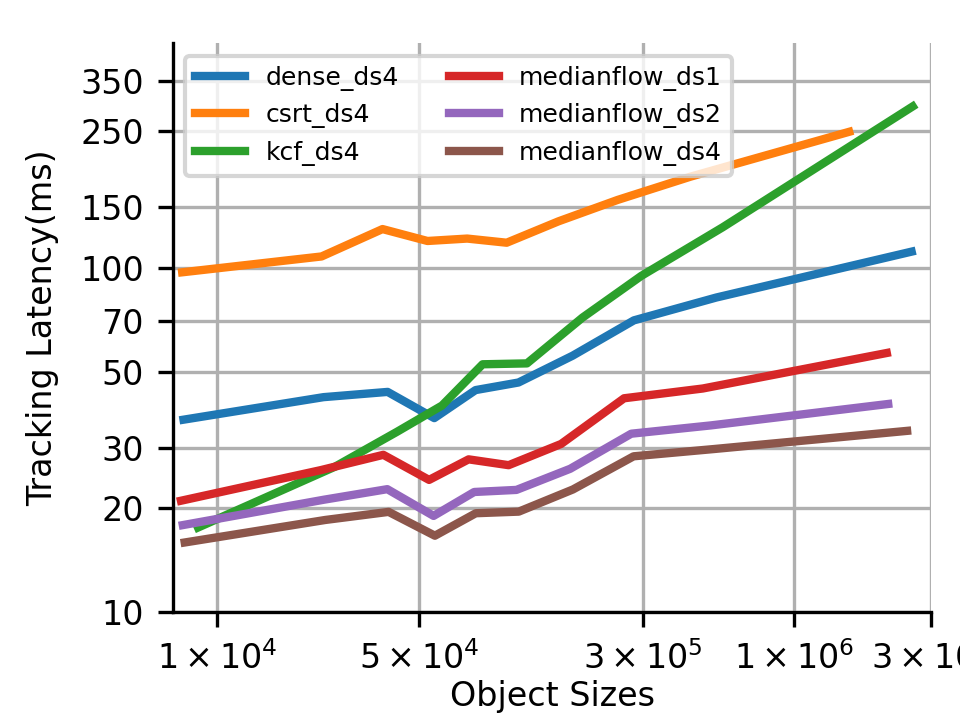}
        \vspace{-1em}
        \caption{The latency curve of object trackers on different sizes of the objects on the validation dataset.}
        \label{fig:tracker_latency_sizes_feature}
        \vspace{-0.5em}
    \end{minipage}
    \hfill
    \begin{minipage}[t]{0.3\linewidth} 
        \includegraphics[width=1.05\linewidth]{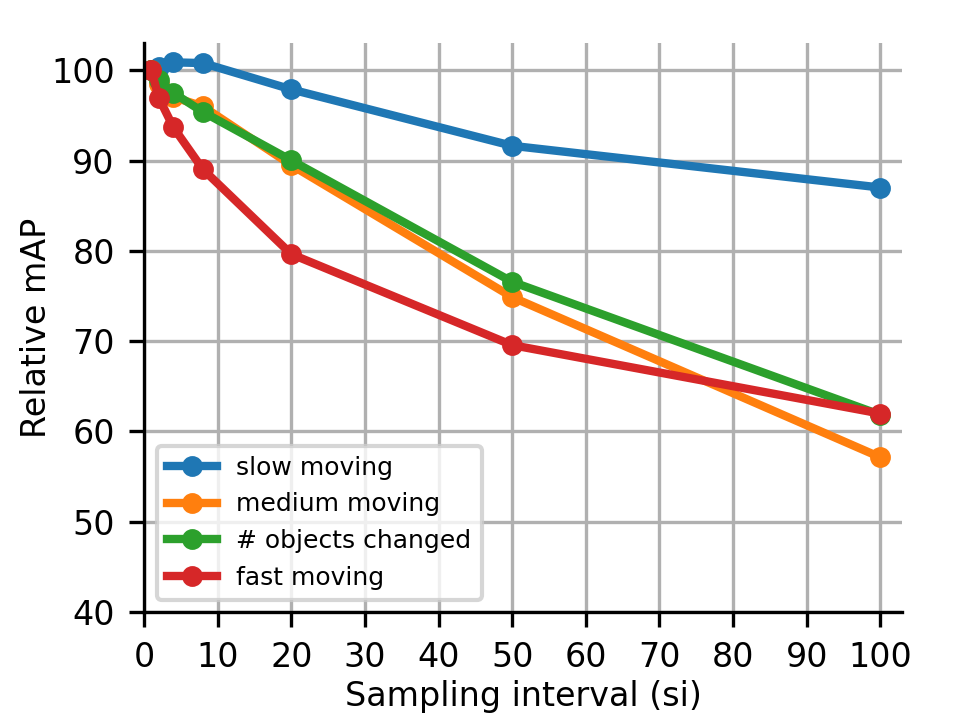}
        \vspace{-1em}
        \caption{Detection plus MedianFlow tracker vis-\`a-vis detection-only branch on slow/medium/fast subsets.}
        \label{fig:movement_ablation}
        \vspace{-0.5em}
    \end{minipage}
    \vspace{-0.5em}
\end{figure*}

We empirically verify that the latency of the object trackers is affected by both the number and sizes of the objects, as shown in Figure~\ref{fig:tracker_latency_nobjs_feature} and~\ref{fig:tracker_latency_sizes_feature}. Specifically, we use 10\% of the ImageNet video object detection~(VID) training dataset to generate the latency data samples. The detailed data split can be found in Section~\ref{sec:datasets}. 
We use all latency data in our validation set to plot Figure~\ref{fig:tracker_latency_nobjs_feature} and~\ref{fig:tracker_latency_sizes_feature}. 
Results have shown that the number of objects ($n\_obj$) and the average size of objects ($avg\_size$) have a significant impact on the tracking latency, which we use as the object's basic features. 

\subsubsection{Object Movement Features}
We use {\em the recent movement of objects} as a feature for modeling the framework accuracy. More rigorously, we define the movement as the Euclidean distance of the objects' centers and we take the mean movement of all the objects in the recent frames.
The intuition is that the faster the objects move in the video frame, the lower the accuracy, especially for the execution branches with higher sampling interval. We use the same data split as in Figure~\ref{fig:tracker_latency_nobjs_feature} and \ref{fig:tracker_latency_sizes_feature} to generate our accuracy data.
We empirically show this in Figure~\ref{fig:movement_ablation}, where we divide the validation dataset into three subsets --- videos with slow, medium, and fast moving objects and show the accuracy reduction (compared to the detection-only branch) as we increase the sampling interval of the object detection kernel. For the detection kernel, we choose 100-proposal, 576-shape branch. The results show that the accuracy of high $si$ branches ($si=100$) does not drop significantly ($\approx$ 10\%) on slow moving videos but reduces ($>$ 30\%) on fast moving videos.

\subsection{Latency Modeling}
\label{subsec:latency_modeling}

Latency prediction models aim to predict the frame-wise latency of each AB for future frames. Denote $L_{fr}$ as the per-frame latency of our adaptive object detection framework. $L_{fr}$ is a function of the DNN based detection latency $L_{DNN}$ and the tracking latency $L_{tracker}$. If object detection DNN runs every $si$ frames (sampling interval), the latency $L_{fr}$ is given by
\begin{equation}
\label{eq_latency}
  L_{fr} = \frac{L_{DNN}}{si}  + L_{tracker},
\end{equation}
We now describe the models of the detection latency $L_{DNN}$ and the tracking latency $L_{tracker}$, respectively.

\subsubsection{Latency Prediction for Object Detection DNN} The latency of the object detection DNN $L_{DNN}$ is jointly determined by the two detector tuning knobs -- the input image size $shape$ and the number of proposals $nprop$. Moreover, considering the input shape of frames may vary in different videos, we add the $height$ and $width$ of the input image as additional features. These features could be ignored if the video source is a video camera (which outputs fixed sized frames). Besides the input shape of video frames, system $contention$ (CPU/GPU usage and memory bandwidth, as detailed in Section~\ref{sec:contention-generator}) will also affect the DNN latency. Thus, the latency equation of the DNN is given by
\begin{equation}
  L_{DNN} = f_{DNN}(nprop, shape, height, width, contention).
\end{equation}
We fit a quadratic regression model for $f_{DNN}$ to characterize the latency of the detection DNN. Once trained, the regression model is evaluated on a subset of the test set (sparsely sampled), where the mean squared error (MSE) between the prediction $\hat{L}_{DNN}$ and the ground-truth $L_{DNN}$ latency are reported. 

\subsubsection{Latency Prediction for Object Trackers}  
As discussed in Section~\ref{sec:object_feats}, the number of objects and average sizes of objects play a major role for the tracking latency. We further construct a model $f_{tracker}$ to characterize the latency of the object tracker under the system contention. Similar to the detection latency model, we also add the $height$ and $width$ of the input image as additional features. Thus, $f_{tracker}$ is given by:
\begin{equation}
  L_{tracker} = f_{tracker}(height, width, n\_obj, avg\_size, contention)
\end{equation}

We fit quadratic regression models to the ground-truth $L_{tracker}$. Moreover, since the model depends on $n\_obj$ and $avg\_{size}$ of the previous frame, we use the previous frame's $n\_obj$ and $avg\_{size}$ to train $L_{tracker}$. After the training process, we compute the predicted $\hat{L}_{tracker}$ and measure the MSE across a subset of the test set.

\subsection{Accuracy Modeling}
\label{subsec:accuracy_modeling}

Accuracy prediction models aim to predict the expectation of the accuracy of each AB for near future frames. Predicting a model's accuracy in the future is a very challenging task. To this end, we start with a baseline, content-agnostic accuracy model that models the average accuracy of approximation branches on an offline validation set. Improving on the baseline, we propose a content-aware model that estimates the accuracy based on the input video content. 

The accuracy of an object detector is usually defined by the metric mean average precision (mAP). However, we find that predicting the absolute mAPs given a test video is difficult. To address this issue, we propose to convert the absolute mAP metric into a relative percentage metric. More precisely, a base branch is identified in the detection framework using the detection-only branch ($si=1$) with $nprop=100$ and $shape=576$. This base branch sets the performance upper-bound for all approximation branches (62.3\% mAP on the validation set). The mAP of each AB is normalized to its percentage value by dividing its mAP by the base branch's mAP.  

\subsubsection{Content-agnostic Accuracy Modeling} 
We first present a baseline model that predicts the average accuracy of each AB on a target dataset, \ie the validation set. Different from the latency models, the factors on the accuracy are coupled all together (\ie no distinction between detection DNN and tracking). Thus, we have a single unified model, given by: 
\begin{equation}
  A=f_{A}(si, shape, nprop, tracker, ds)
\end{equation}
where $tracker$ is the tracker type and $ds$ is the downsampling ratio of the input to the tracker. A decision tree model $f_{A}$ was learned to predict the accuracy $A$, trained with the MSE loss across the whole training dataset. 

\subsubsection{Content-aware Accuracy Model} 
A content-agnostic accuracy model is independent of the video content, and thus often fails to predict the accuracy of our detection framework on individual videos that may differ significantly from the average accuracy on the validation set. We further design a content-aware accuracy model that predicts the accuracy of an AB taking the video characteristics into account. Our model is presented in Eq.\ \ref{eq_af} where both the configurations of the detection framework and content characteristics are taken into consideration.
\begin{equation} 
  \label{eq_af}
  A = f(si, shape, nprop, tracker, ds, movement)
\end{equation}
where $movement$ is the object movement features extracted from the video content. 
We find that linear regression models demonstrate 8\% improvement (in MSE) vis-\`a-vis the decision tree models used in our content-agnostic models. 

\vspace{-0.5em}
\subsection{Synthetic Contention Generator}
\label{sec:contention-generator}

Synthetic Contention Generator (CG) is designed as a stand-in for any resource contention on the device that may affect \name.
A detection framework may suffer from unpredictable levels of resource contention when it is running on mobile platforms due to the instantiation of other co-located applications, for which we will not have information.
We focus on three important types of resources on mobile platforms --- CPU, memory bandwidth (MB), and GPU. We control CPU contention by the number of CPU cores our CG occupies. 
We control MB contention by the amount of memory-to-cache bandwidth that it consumes. The code is modified from the widely used STREAM benchmark~\cite{McCalpin2007, McCalpin1995} that is meant to measure the MB capacity of a chip. For the GPU contention, we control the number of GPU cores that are utilized. The three-dimensional CG is orthogonal, which means we can tune each dimension without affecting the other dimensions.
The CG is representative because we executed and mapped the contention caused by some widely-used applications in the 3D contention space (Table~\ref{table:3d_annotation_real_apps}). The first one is an anomaly detection program that uses Robust Random Cut Forest (RRCF) \cite{guha2016robust} to detect anomalies from a local temperature and humidity sensor data. We also used our two object detection DNNs, Faster R-CNN and YOLOv3, for checking how much contention they generate.

\begin{table}[h]
  \centering
  \caption{Applications running in the 3D contention space.}
  \label{table:3d_annotation_real_apps}
  \scalebox{0.9}{
  \begin{tabular}{|p{1in}|p{0.5in}|p{0.7in}|p{0.5in}|}
     \hline
     Real Apps & CPU & MB (MB/s) & GPU \\
     \hline
     Anomaly detection & 99.80\% & 500 & 0\% \\
     \hline
     Faster R-CNN & 69.75\% & 1000 & 99\%  \\
     \hline
     YOLOv3 & 65.85\% & 800 & 98.50\%  \\
     \hline
  \end{tabular}
  }
\end{table}

\vspace{-0.5em}
\subsection{Profiling Cost and Sub-sampling}
\label{sec:profiling_cost}

The cost of collecting ground truth data with design features for performance prediction models is significant without proper sampling techniques. We measure our profiling cost for the accuracy, detection latency, and tracker latency models in Table~\ref{table:cost_profiling}.

\begin{table}[t]
  \centering
  \caption{Cost of profiling.}
  \label{table:cost_profiling}
  \vspace{-1em}
  \scalebox{0.9}{
      \begin{tabular}{|p{1.1in}|p{2.2in}|}
         \hline
         Task                                        & Cost \\
         \hline
         Framework accuracy                          & 2,414 hr $\cdot$ core (20\% of the configurations) \\
         \hline
         Detection latency                           & 7 hr $\cdot$ machine w/ 15 out 1 million sampling  \\
         Tracker latency                             & 1 hr $\cdot$ machine w/ 169 out 1 million sampling  \\
         \hline
      \end{tabular}
  }
  \vspace{-1em}
\end{table}

To efficiently collect the profiling data, we use the master and worker model, where the master node manages a list of configurations of the detection framework and distributes the profiling work, while workers run the particular configuration to collect the training data for the modeling. As the feature space is huge, we sparsely sample the multi-dimensional space of (``number of proposals'', ``resized shape'', ``sampling interval'', ``tracker'', ``down-sampling ratio of the tracker'). We finally use 20\% of the configurations to train our accuracy model.

Similar sub-subsampling techniques are used for the latency models as well, and we sample data points on videos of various height and width, various numbers of objects and object sizes, under discrete 3D contention levels. We finally use 15 out of a million feature points (defined in Section~\ref{sec:implementation_CG} and ~\ref{sec:implementation_values})  to train our detection latency model and 169 out of a million feature points to train our tracker latency model.

\vspace{-0.5em}
\subsection{Scheduler}
\label{subsec_scheduler}

The scheduler is the core component of \name that makes the decision at runtime on which AB should be used to run the inference on the input video frames. Formally, the scheduler maximizes the estimated detection accuracy of \name given a latency requirement $L_{req}$. This is done by identifying a feasible set of branches that satisfy the target latency requirements, and choosing the most accurate branch. In case of an empty feasible set, the fastest branch is returned. Thus, we formulate the optimal AB $b_{opt}$ as follows,

\begin{equation}
  \centering
  b_{opt} = \begin{cases} 
    \argmax_{b \in \hat{B}} (A_{b}),\ &\text{if}\ \hat{B} \neq \emptyset,  \\
    \argmin_{b \in \mathcal{B}} (L_{est,b}) & \text{otherwise}
  \end{cases}
  \label{eq:opt-latency-constraint}
\end{equation}

where $\mathcal{\hat{B}}$ is all ABs considered, $\hat{B}$ is the feasible set, \ie $\hat{B} = \{b \in B\} \ \ \text{iff} \ \ L_{est,b} < L_{req} $, $A_{b}$ and $L_{est,b}$ are the estimated accuracy and latency of the AB respectively. The search space $\mathcal{\hat{B}}$, composed of five orthogonal knobs, has millions of states. To reduce the scheduler overhead, we use a sampling technique in Section~\ref{sec:implementation_values} and design light-weight online feature extractors.

To further reduce the scheduler overhead and enhance our system robustness, we restrict the scheduler to make decision at least every $sw$ frames. The motivation of introducing $sw$ is to prevent the scheduler to make very frequent decisions. Specifically, we set $sw = max(8, si)$. The scheduler will thus make a decision at least every 8 frames. When the scheduler chooses a branch with a long $si$, it will make a following decision every $si$ frames. In addition to the latency of the detection and tracking kernels, we add switching overhead $L_{sw}$ and the scheduler overhead $L_{sc}$ into the overall latency estimation of a AB $b$, \ie $L_{est,b} = L_{b,fr}+(L_{sw}+L_{sc})/{si}$. We also design the light-weight online feature extractors so that we can adapt seamlessly to the content and contention changes.

\noindent \textbf{Online content feature extractor}
The online content feature extractor maintains the content features of the video by extracting $height$, $width$ from current frame, memorizing $n\_obj$, $avg\_size$ of last frame and $movement$ from past frames. It is lightweight in terms of the compute load it puts on the target platform and this is desirable since we have to extract the features at runtime on the target board for feeding into our models.

\noindent \textbf{Online contention sensor}
The online contention sensor is designed to sense the contention level in the system so that we can refer to the modeling and make the right prediction on the latency of each AB. There are generally two approaches---one is to probe the system by directly measuring CPU, memory bandwidth and GPU usage by other processes, and the other is to record the latency of \name in the past few runs and match with the offline log. Although the first one can theoretically get the ground truth of the resource contention, it is not practical. As a normal application in the user space, it is difficult for \name to collect the exact resource information from other processes. The hardware is also lacking sufficient support for such fine-grained measurement on mobile or embedded devices~\cite{tancreti2011aveksha}. In contrary, the offline latency log under various contention levels and the online latency log of the current branch in the past few runs are a natural observation of the contention level. Thus, we proposed the log-based contention sensor. 

The log-based contention sensor tries to find a contention level where the offline latency log matches the averaged online latency most closely. We use the nearest-neighbor principle to search for such contention levels in our pre-defined orthogonal 3D contention space. As multiple contention levels may cause the same impact on the latency of a given AB, we call it a cluster of contention levels and we pick one level out of it as the representative.
In comparison to some previous work in the systems community~\cite{fang2018nestdnn}, our contention sensor is lightweight, efficient, and does not require additional privileges at system level, making it a more practical offering in real-world systems.

\vspace{-0.5em}
\section{Implementation}
\label{sec:implementation}

We implement \name in Python 3 and C with tensorflow-gpu, CUDA, and cuDNN libraries and release our code at \url{https://github.com/StarsThu2016/ApproxDet}. For detection kernel, we choose Faster R-CNN due to its high accuracy and moderate computational burden. For tracking kernel, we implement four variants and introduce the details in Section~\ref{sec:implementation_values}.

\vspace{-0.5em}
\subsection{Configuration of the Tuning Knobs}
\label{sec:implementation_values}

Our five tuning knobs include the sampling interval ($si$), the input image size ($shape$) to the detection DNN, the number of proposals ($nprop$) in the detection DNN, the type of object tracker ($tracker$) and the downsampling ratio of the input to the tracker ($ds$). We now describe the implementation details of these knobs, including their data types and value ranges.

\noindent \textbf{Sampling Interval ($si$)}. $si$ defines the interval of running the object detector. The object tracker runs on the following $si-1$ frames. For example, our system runs object detection on every frame when $si=1$. 
To reduce the search space of $si$, we constrain $si$ in a preset set---$\{1,2,4,8,20,50,100\}$. These pre-defined $si$ are chosen empirically to cover common video object detection scenarios. With the max value of $si=100$, the detector runs at a large interval of 3-4 seconds and the tracker runs in-between. 

\noindent \textbf{Input Video Frame Shape to Detector ($shape$)}. $shape$ defines the shortest side of the input video frame to the object detector. The value of $shape$ must be a multiple of 16 to make the precise alignment of the image pixels and the feature map~\cite{ren2015faster}. We set the $shape$ range from 224 to 576, since smaller shape than 224 significantly reduces the accuracy and larger shape than 576 will result in heavy computational burden and does not improve the accuracy based on results on the validation set.

\noindent \textbf{Number of Proposals ($nprop$)}. $nprop$ controls the number of candidate regions considered for classification in the object detector. We limit the value of $nprop$ (integer) between 1 and 100. With $nprop=1$, only the top ranked proposal from RPN is used for detection. Increasing $nprop$ will boost the detector's performance yet with increased computational cost and runtime. 

\noindent \textbf{Type of Trackers ($tracker$)}. $tracker$ defines which tracker to use from MedianFlow~\cite{MedianFlow}, KCF~\cite{KCF}, CSRT~\cite{CSRT}, and dense optical flow trackers~\cite{farneback2003two}. These trackers are selected based on their efficiency and accuracy. Different trackers have varying performance under different scenarios. For example, CSRT tracker is most accurate among these trackers, but is also most time consuming. MedianFlow tracker is fast and accurate when a object move slowly in the video, yet have poor performance for a fast moving object. We use the implementation from OpenCV for all trackers.

\noindent \textbf{Downsampling ratio for the tracker ($ds$)}. $ds$ controls the input image size to the tracker. The value of $ds$ is limited to 1, 2, and 4, \ie no downsampling, dowsampling by a factor of 2 and 4, respectively. A larger $ds$ reduces the computational cost, and favors the tracking of fast moving objects. A smaller $ds$ increase the latency, yet provide more accurate tracking of slowly moving objects. 

\vspace{-0.7em}
\subsection{3D Contention Generator (CG)}
\label{sec:implementation_CG}

\noindent Our 3D CG is lightweight in code size. It is configured to generate contention in CPU, memory bandwidth (MB), and GPU (as introduced in Section~\ref{sec:contention-generator}). For MB CG, we modify STREAM by having the code write continuously to a 152 MB memory space and controlling the interval of array elements to operate on the array data so as to control the MB occupied. We add a feedback loop to dynamically adjust the number of elements in the array to be skipped for the write operation, thus maintaining the MB contention at the experimentally given level. To increase the maximum contention that can be generated by the MB CG, we spread out the MB CG among the CPU cores on which contention is to be generated and in aggregate can generate an intense bandwidth contention of up to 18 GB/s when 6 CPU can be used. The maximum input for the MB CG depends on both the CPU and MB part, the more CPU we can occupy, the higher maximum MB contention we can achieve. For the GPU CG, we fixed the working frequency of TX2 board at 1300 MHz. Then our GPU CG performs add operation on a certain size of arrays by using a CUDA program. By changing the size of the arrays and the input to the CUDA kernel functions, we control the number of GPU cores that are kept busy. We generate contention from 1\% to 99\%, as measured by \texttt{tegrastats}. For the experiments, we use 11 discrete levels 1\%, 99\%, and 9 levels in increments of 10\% starting at 10\%. If there is no MB or GPU contention specified and we need CPU contention on a certain number of CPU cores, we use a MB CG with minimum input 1MB/s to serve as the CPU CG and pin it to the experimentally given number of cores. For the experiments, we use 6 discrete levels from 100\% to 600\% in increments of 100\%. 

\vspace{-0.8em}
\subsection{Training of Latency and Accuracy Models}

The latency and accuracy model in \name is trained using a subset of the validation set of ImageNet VID. Specifically, we sparsely sample the 5-D feature space $(si, shape, nprop, tracker, ds)$ and run \name using the sampled configuration on a subset of videos randomly sampled from the validation set. Standard gradient descent is then used for fitting the regression. And CART is used for the decision tree. To train content-aware accuracy model, during the training phase, the $movement$ feature as the average motion across all objects and all frames is required in each video snippet. During the test phase, since movement of the objects are not available, we use average across all past frames as a substitute.
\section{Evaluation}
\label{sec_evaluation}

\subsection{Evaluation Platform}
We evaluate \name on an NVIDIA Jetson TX2 board~\cite{tx2}, which includes 256 NVIDIA Pascal CUDA cores, a dual-core Denver CPU, a quad-core ARM CPU on a 8GB unified memory between CPU and GPU. The specification of this board is comparable to what is available in today's high-end smartphones such as Samsung Galaxy S20 and Apple iPhone 11 Pro.
We train our neural network models on a server with NVIDIA Tesla K40c GPU with 12GB dedicated memory and an octa-core Intel i7-2600 CPU with 24GB RAM. 
For both the TX2 and the edge server, we install Ubuntu OS and Tensorflow v1.14, Pytorch v1.1, and MXNet v1.4.1.

\vspace{-0.5em}
\subsection{Datasets, Task, and Metrics}
\label{sec:datasets}

We evaluate \name on the object detection task using ILSVRC 2015 VID dataset~\cite{ILSVRC15}. For the purpose of training, since ILSVRC 2015 VID training set is a video dataset, due to the redundant video dataset and limited resources, we follow the practice in~\cite{kang2017t} such that the VID training dataset is sub-sampled every 100 frames. We use 90\% of this video dataset as training set to train \name's DNN model and keep aside another 10\% as validation set to fine-tune \name (modeling). To evaluate \name's system performance, we use ILSVRC 2015 VID validation set -- we refer to this as the ``test set'' throughout the paper. For all our baselines, we follow the data split of \name's DNN model to train and test. 

We use latency and mean average precision (mAP) as the two metrics. We define the latency as the short-window averaged latency among one detection frame and its following tracking frames. The definition applies to \name's baselines with different trackers used. The latency also includes the overheads of the respective solutions, e.g., the switching overhead, the execution time of the online feature extractor, the online contention sensor, and the scheduler. For common detection methods, they will use low confidence threshold, e.g., 0.001~\cite{ren2015faster,liu2016ssd} to achieve higher mAP. However, low confidence threshold will lead to many false positive outputs, which is not practical in real-world systems. To simulate the environment of real-world systems, we use a high confidence threshold of 0.3 to reduce the number of output objects of detection DNN, and make it the same for all baseline detection DNNs.
Note that we always \textit{use the ground truth annotations} available in the datatset to examine the \textit{true} accuracy and never use the detection results of some state-of-the-art models as pseudo ground truth (fake annotations as in~\cite{hsieh2018focus}). 

\vspace{-0.5em}
\subsection{Baselines}
\label{sec:baselines}

In this section, we will introduce baseline models used in our paper.

\vspace{0.2em}
\noindent \textbf{Faster R-CNN (FRCNN)}~\cite{ren2015faster} is a popular two-stage detector in the computer vision community. In this paper, we vary the $nprop$ and $shape$ of \name's detection DNN to create different FRCNN baseline detectors. From this, we get a total of 28 ($4 shape \times 7 nprop$) FRCNN baseline models. In these baseline models, we follow the multi-model design on the detection model, where model variants in different sizes (number of proposals and resized shape) achieve different latency and accuracy specifications. We also profile the latency of these model variants and evaluate FRCNN with our scheduler defined in Section~\ref{subsec_scheduler}.

\vspace{0.2em}
\noindent \textbf{FRCNN+MedianFlow (FRCNN+MF):}
Since FRCNN only uses detection, we want to enhance this baseline for a fair comparison. For the enhanced baseline, we follow the mainstream ``detection plus tracking'' design in lightweight object detection tasks. We pick an unmodified detection variant with the highest accuracy~($nprop$ = 100, $shape$ = 576) and a fast object tracker---MedianFlow (without downsampling technique). Thus, the enhanced baseline models include FRCNN plus MedianFlow tracker with varying $si$. We also profile the latency of these model variants (each $si$) and evaluate FRCNN+MF with our scheduler defined in Section~\ref{subsec_scheduler}. To reduce the scheduler cost, we perform the same sampling strategy in Section~\ref{sec:implementation_values}. Though we can pick different models based on the latency budget, these models are static and cannot respond to contention and context changes.

\vspace{0.2em}
\noindent \textbf{AdaScale:} Among latest methods, we choose AdaScale~\cite{chin2019adascale} as it dynamically adjusts the input scale to improve accuracy {\em and} running speed simultaneously. AdaScale is thus most relevant to our work, with similar tuning knobs for resource-constrained devices.
For conducting a fair comparison between AdaScale and our proposed \name, we re-implemented AdaScale and modified the following settings:

\begin{itemize}
    \item Pretrained settings: In AdaScale, they use pretrained models on ImageNet VID + ImageNet Detection (DET). In this paper, \name and AdaScale both start from ImageNet pretrained models.
    \item Training sets: In AdaScale, they use ImageNet detection dataset~(DET) joint with ImageNet VID dataset and subsample every 15 frames in each VID video. In this paper, we remove all training images in ImageNet DET dataset and we use same sampling techniques~(sub-sampled every 100 frames) for training Adascale.
\end{itemize}

\vspace{0.2em}
\noindent \textbf{YOLOv3:} 
YOLOv3 is a modern one-stage detector with a fast running speed. It is widely used in many mobile applications. In this paper, we use the same training/testing set and scheduler as \name to train and evaluate YOLOv3.

\begin{figure}[th]
    \begin{minipage}[t]{0.48\linewidth} 
        \centering
        \includegraphics[width=1.05\columnwidth]{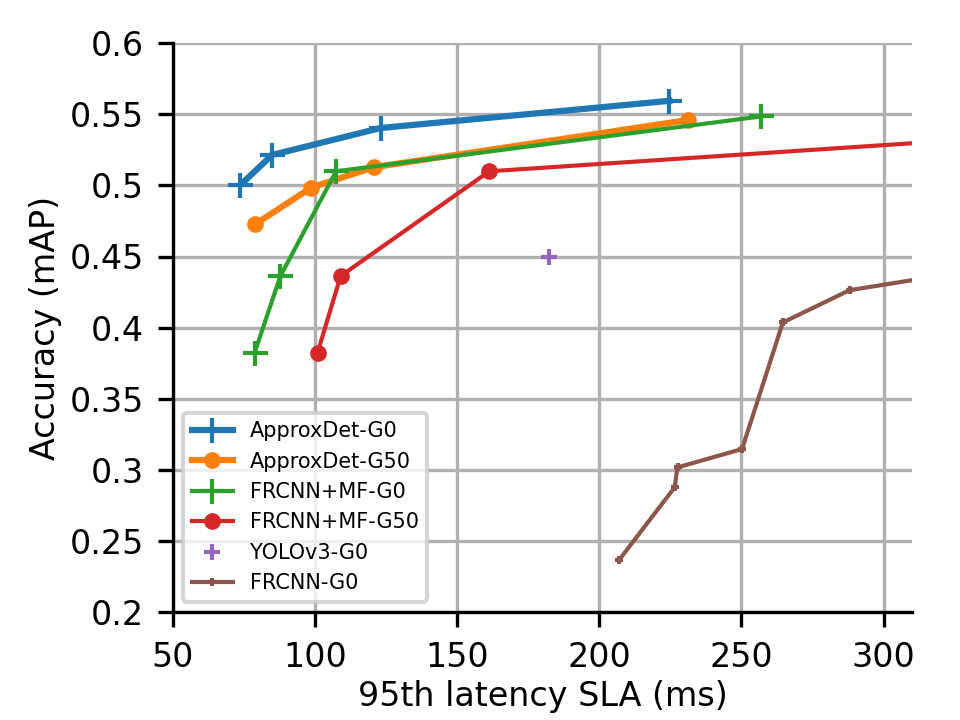}
        \vspace{-0.5em}
        \caption{Accuracy of the models vs 95-th latency SLA. G50 represents the 50\% GPU contention and G0 represents no contention. (zoomed in)}
        \label{fig:accuracy_w_contention_micro}
        \vspace{-0.5em}
    \end{minipage}
    \hfill
    \begin{minipage}[t]{0.48\linewidth} 
        \centering
        \includegraphics[width=1.05\columnwidth]{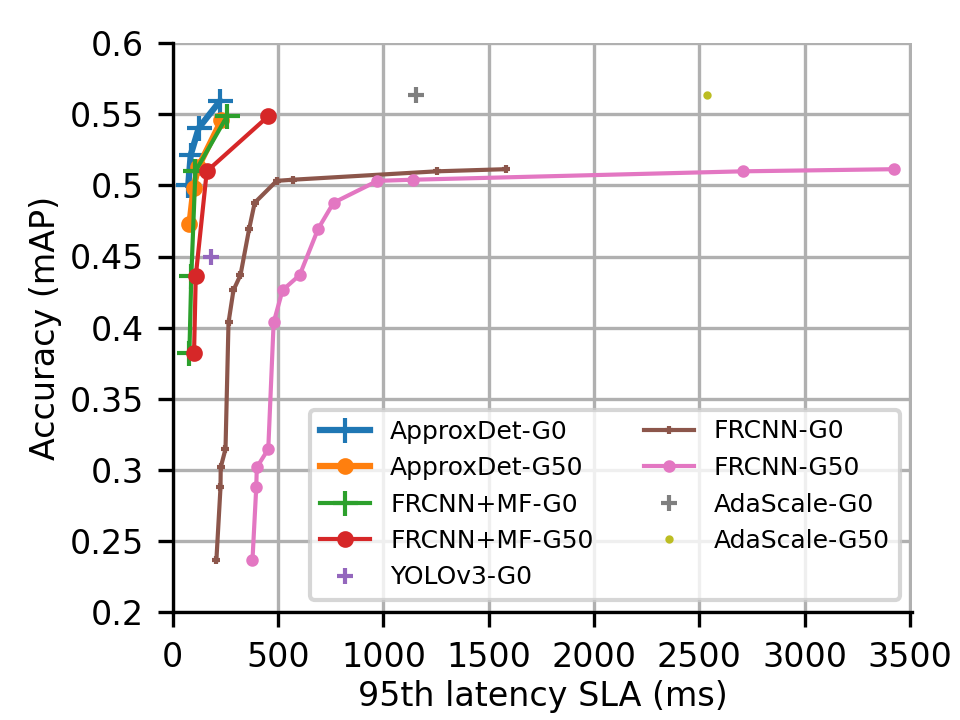}
        \vspace{-0.5em}
        \caption{Accuracy of the models vs 95-th latency SLA. g50 represents the 50\% GPU contention and g0 represents no contention. (zoomed out)}
        \label{fig:accuracy_w_contention_macro}
        \vspace{-0.5em}
    \end{minipage}
    \vspace{-1em}
\end{figure}

\subsection{End-to-End Evaluation on Budgeted Latency}

We first examine the end-to-end performance of \name vs. various baselines. Figure~\ref{fig:accuracy_w_contention_micro} shows a micro-view of the accuracy and 95-th percentile latency~\footnote{We choose the 95-th percentile latency service level agreement (SLA) because it is a promise to the users that in most cases the latency is below the number and shows much stronger latency guarantee than either mean or median latency.} under no contention (G0) and injected 50\% GPU contention (G50) from the CG. The reason why our evaluation injects GPU contention only, is that the inference time of the DNN has the most impact on the detection latency and such inferences mostly run on the device's GPU (when it is available). Hence, the impact of GPU contention would focus on the most interesting scenarios showing how resilient \name is with respect to changes in the dynamics within the mobile device.

The results show the superior accuracy-vs-latency trade-off (blue vs. green vs. purple) over a FRCNN+MF and YOLOv3 under no contention where particularly, \name's accuracy is 11.8\%, 9.5\% higher than FRCNN+MF given 80ms and 90ms latency SLA and also 9.1\% higher than YOLOv3 given 170ms latency SLA (exactly following YOLOv3's posterior 95-th latency). The accuracy gain over FRCNN+MF reduces as latency SLA grows larger since both \name and FRCNN+MF will merge into detection-only (or detection in every alternate frame) branch. 

Furthermore, when 50\% GPU contention is injected, \name with strong adaptability on sensing and reacting to the contention, the accuracy drops by around 2\% and preserves the 95th latency SLA with minor changes (max 16\%). However, the best baseline FRCNN+MF cannot adapt to the contention and the latency increases significantly (25\% - 50\%). 
Figure~\ref{fig:accuracy_w_contention_macro} shows the performance of FRCNN and AdaScale. Both of their latency increase by 75\% to 120\% with increased contention levels. This is significantly worse than \name. Further, FRCNN is inferior to FRCNN+MF in terms of accuracy-latency tradeoff. This is because adding the light-weight MedianFlow tracker largely reduces the latency while preserving most of the accuracy.

\begin{figure}[th]
    \begin{minipage}[t]{0.48\linewidth} 
        \centering
        \includegraphics[width=1.05\columnwidth]{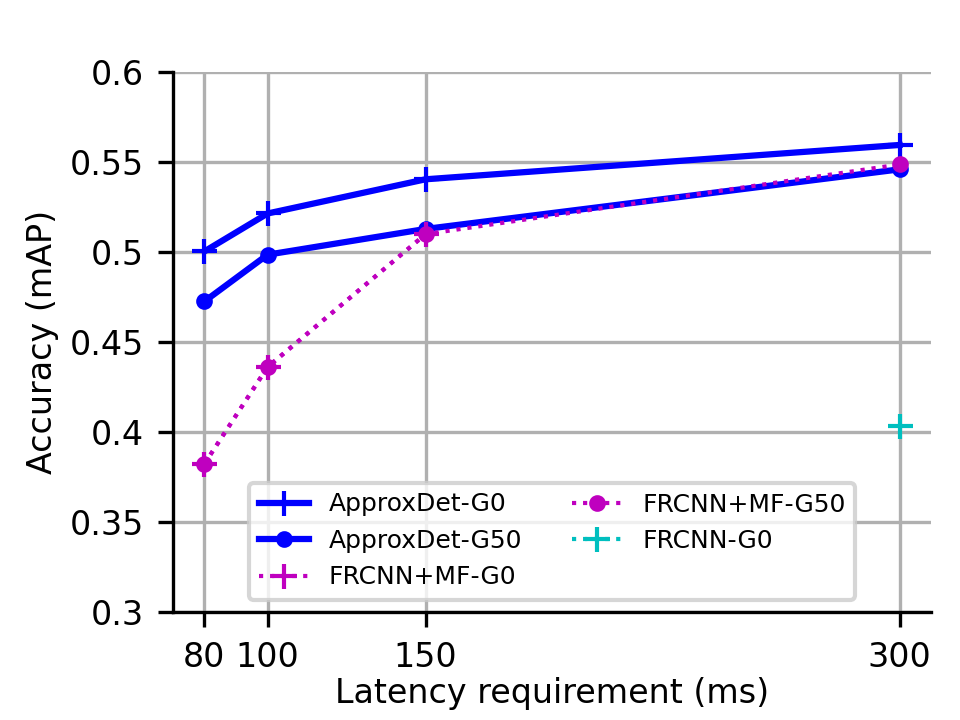}
        \vspace{-0.5em}
        \caption{Given latency requirement, accuracy (mAP). (FRCNN+MF)'s accuracy does not change since it cannot adapt to contention and the execution is the same. FRCNN does not work for the latency SLA $<$ 300 ms.}
        \label{fig:end_to_end_dyn_models_accuracy}
        \vspace{-0.5em}
    \end{minipage}
    \hfill 
    \begin{minipage}[t]{0.48\linewidth} 
        \centering
        \includegraphics[width=1.05\columnwidth]{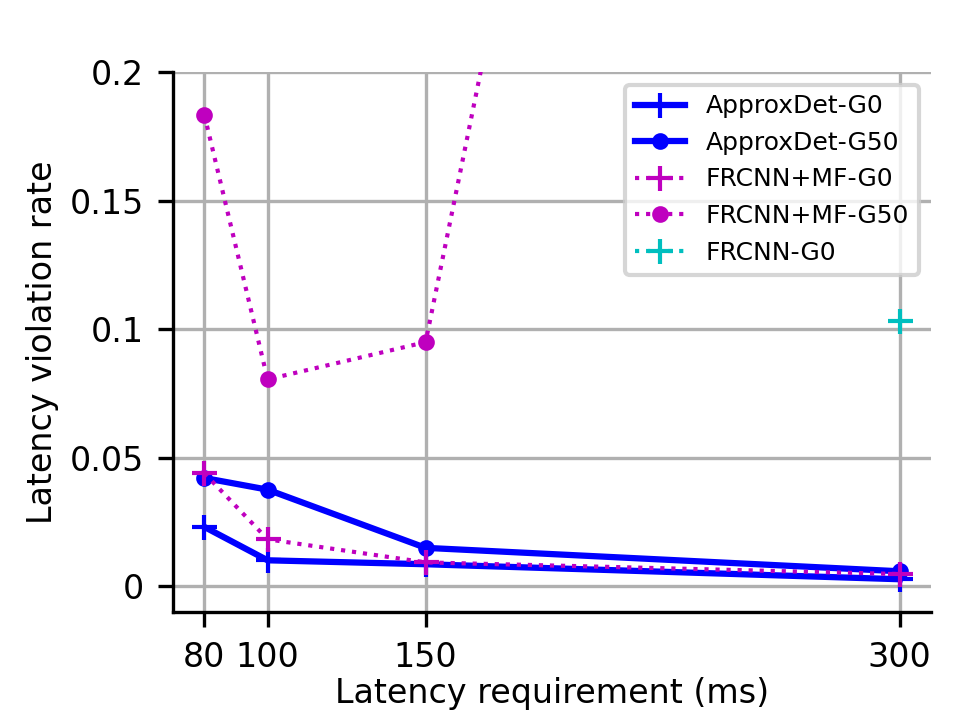}
        \vspace{-0.5em}
        \caption{Given latency requirement, latency violations. The latency violation is higher with contention although FRCNN+MF is using a very conservative p95 scheduler. }
        \label{fig:end_to_end_dyn_models_latency_violation}
        \vspace{-0.5em}
    \end{minipage}
    \vspace{-1em}
\end{figure}

Although accuracy vs latency SLA shows the posterior performance metric of each model, we also want to examine under a certain latency requirement, how each solution chooses a model variant to execute and what the accuracy (Figure~\ref{fig:end_to_end_dyn_models_accuracy}) and the latency violation rates (Figure ~\ref{fig:end_to_end_dyn_models_latency_violation}) are. We evaluate at 80ms, 100ms, 150ms, and 300ms. Latency requirement that is smaller than 80ms is not chosen because no branch will be returned from both baselines and larger latency requirement is not meaningful as discussed before. Figure~\ref{fig:end_to_end_dyn_models_accuracy} and ~\ref{fig:end_to_end_dyn_models_latency_violation} show that under no contention scenario (GO), \name and FRCNN+MF are equally good at controlling low latency violations (\name is slightly smaller), while the accuracy of \name is 11.8\%, 8.5\%, 3.0\%, and 1.1\% higher than FRCNN+MF. Even under great reduction to 80ms latency requirement, \name is still able to maintain the accuracy above 50\%. 

Then, as 50\% GPU contention is injected, the accuracy of \name is slightly reduced (by 2\%) and is as good as FRCNN+MF under 150ms and 300ms latency requirements. However, \name is still able to control within the 5\% violation rate while FRCNN+MF has a violation rate of 18.3\%, 8.0\%, 9.5\%, and 100\%. To summarize, \name is best at reducing the latency requirement to as low as 80ms with slightly reduced accuracy and is able to adapt well to the contention without violating the latency requirements. 

\subsection{Case Studies on Changing Conditions}
Although the macro benchmark shows the overall performance of \name on a whole dataset, it is mostly the static behavior of the models since the latency requirement and resource contention levels do not change. To examine how \name adapts to the runtime changing conditions, we set up these case studies by injecting changing contention levels and latency requirements.

\begin{figure*}[th]
    \begin{minipage}[t]{0.31\linewidth} 
        \centering
        \includegraphics[width=1\columnwidth]{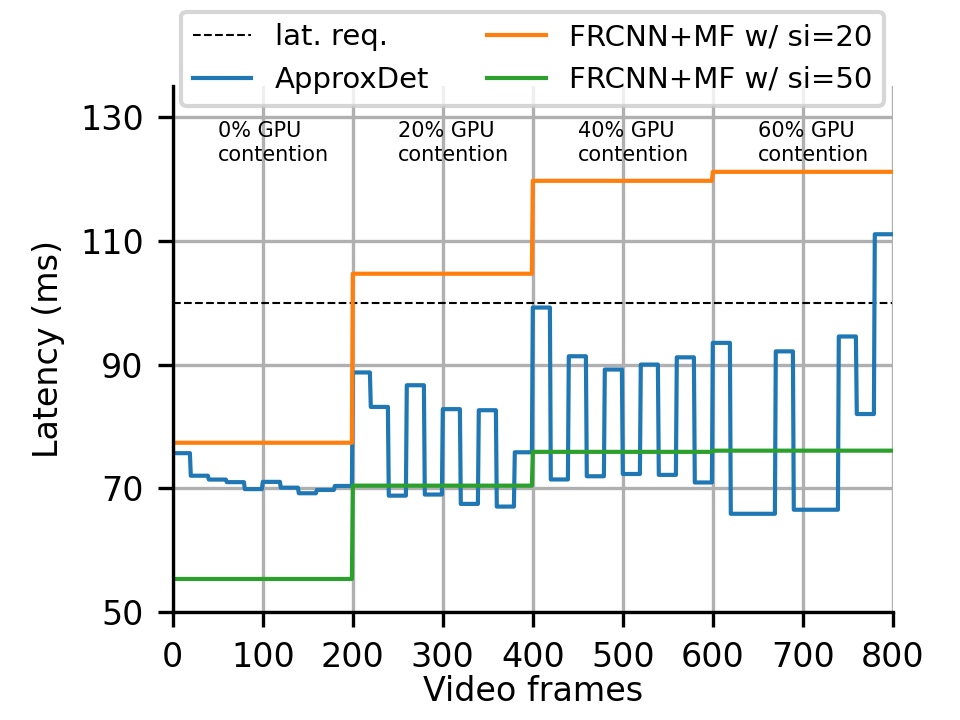}
        \vspace{-1em}
        \caption{Models' latency under changing contention.}
        \label{fig:result-changing-contention}
        \vspace{-0.5em}
    \end{minipage}
    \hfill
    \begin{minipage}[t]{0.31\linewidth} 
        \centering
        \includegraphics[width=1\columnwidth]{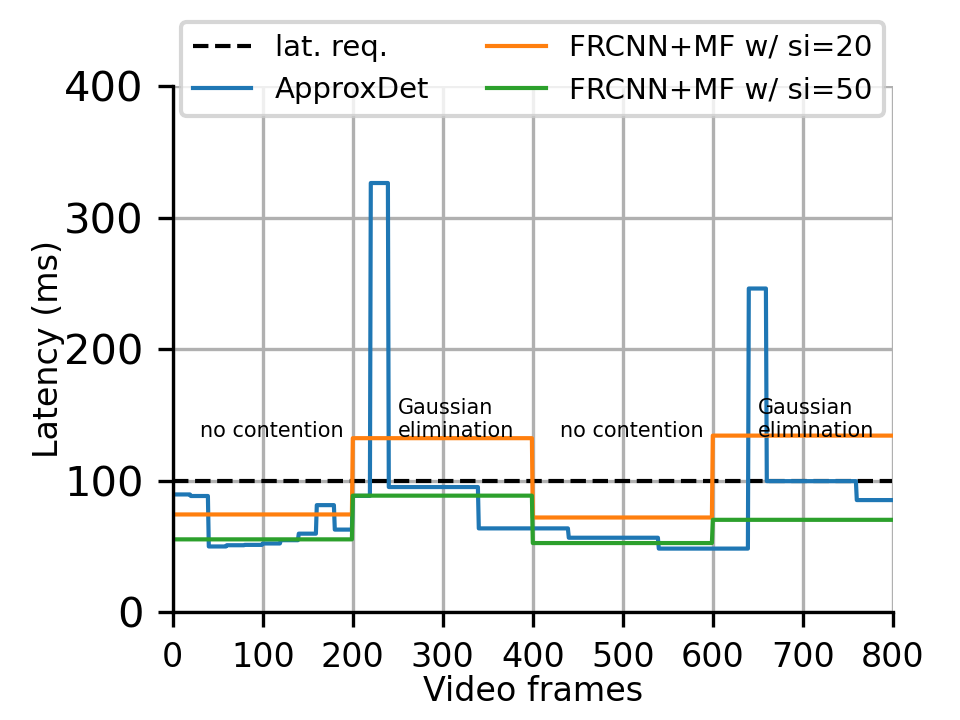}
        \vspace{-1em}
        \caption{Models' latency without and with real app.}
        \label{fig:result-changing-contention-real-app}
        \vspace{-0.5em}
    \end{minipage}
    \hfill
    \begin{minipage}[t]{0.31\linewidth} 
        \centering
        \includegraphics[width=1\columnwidth]{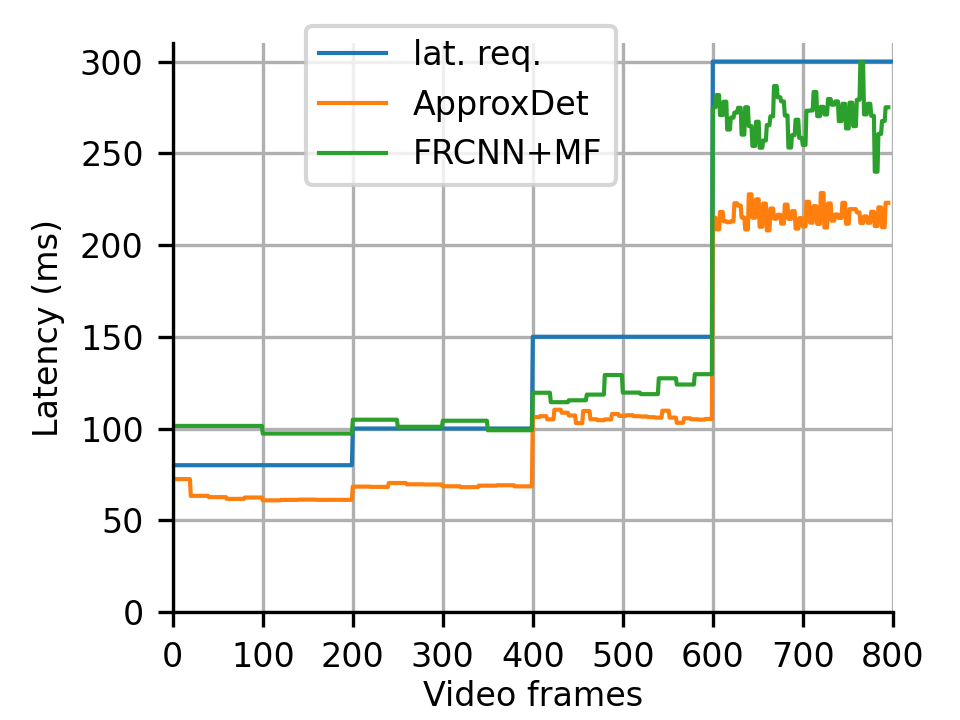}
        \vspace{-1em}
        \caption{Models' latency under changing latency budget.}
        \label{fig:temporal_latency_changing_latreq}
        \vspace{-0.5em}
    \end{minipage}
\end{figure*}

\vspace{0.2em}
\noindent \textbf{Changing Contention Levels}.

We first examine the performance of \name given a fixed latency requirement of 100 millisecond on one test video. We then manually inject the GPU contention through CG and gradually tune up the contention level by 20\% for every 200 frames. Figure~\ref{fig:result-changing-contention} shows that \name with quick sensing and adaptation, is always able to meet the latency requirements while a FRCNN+MF baseline will either exceed the latency requirement by 20\% ($si=20$) or stay too conservative ($si=50$) and suffers from low accuracy (there is a 7\% mAP difference between the two branches on the test dataset). 

In addition, we also pick a real application---Gaussian Elimination from the Rodinia Benchmark Suite~\cite{che2009rodinia}, a GPU-intensive linear algebra routine widely used by many applications. We examine the performance of \name and baselines without and with the background app. Figure~\ref{fig:result-changing-contention-real-app} shows that \name can adapt to the resource contention produced by the background app. The latency goes up dramatically when the app starts running and quickly drops as \name senses such contention and schedule a more efficient AB. A repeated experiment during the 600---800 frames has further confirmed our adaptability. In contrast, the baseline FRCNN+MF with different $si$ are either too conservative or too aggressive under varying contention levels in terms of latency. When the contention app starts running, there is a larger latency spike of \name than FRCNN+MF. This is because our system schedules the AB with smaller $si$ as long as the latency of such AB is below the latency requirement. However, since the object detector takes a larger portion in latency and is more sensitive to GPU contention, the latency increase is much higher when the system has not responded yet.

\vspace{0.2em}
\noindent \textbf{Changing Latency Requirements}.
We then examine the performance of \name given more relaxed latency requirements of 80ms, 100ms, 150ms, and 300ms per frame in four equally chunked phases of 200 frames. Figure~\ref{fig:temporal_latency_changing_latreq} shows that we can always keep up with the latency requirement and always run below the latency requirement while FRCNN+MF, although is choosing a branch that satisfy 95\% of the video frames in the validation dataset, still violates the latency requirement by 20\% under 80ms requirement and is slightly above the threshold given 100ms latency requirement.

\subsection{Micro-benchmark: Performance Prediction Models}

\begin{figure}[th]
    \centering
    \includegraphics[width=0.7\columnwidth]{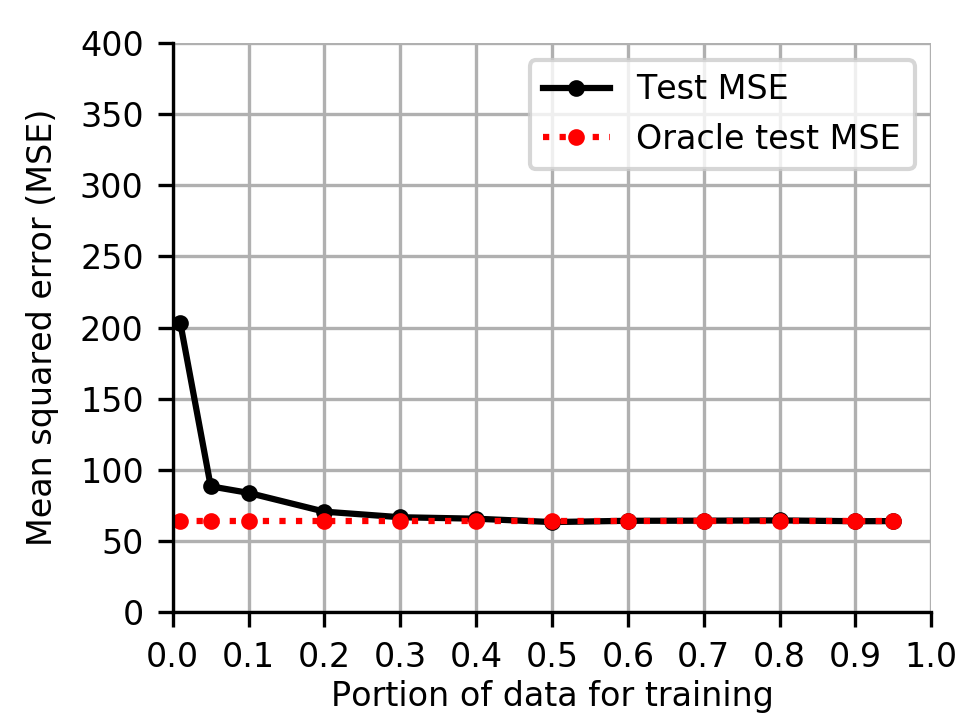}
    \vspace{-1em}
    \caption{MSE of the accuracy prediction model given certain amount of training data in the validation dataset.}
    \label{fig:acc_training}
    \vspace{-0.5em}
\end{figure}

\subsubsection{Accuracy prediction model}
We set up the \textbf{oracle} method as using the ground truth validation data to predict on the test dataset. According to our belief that the accuracy reduction should be same in validation dataset and test dataset, the oracle approach should achieve zero MSE and if it is not zero, it represents the gap of accuracy reduction between the two datasets.

Figure~\ref{fig:acc_training} shows the training curve of the accuracy prediction model. We use different amounts of data to train the model and examine the mean squared error (MSE) on the rest of validation dataset for cross validation and the whole test dataset to report final performance. We can see that with only 20\% of the training data, we are able to predict on the test dataset with 74.58 MSE. We can further reduce the MSE to 71.67 if 95\% of the training data is available, while the oracle predicts with a comparable 71.65 MSE.

\begin{table}[th]
  \centering
  \caption{Precision of the tracker latency prediction models on the validation dataset. Please note that we only show the results of tracking frames on test datasets. In this table, ``no'' means no contention, ``g50'' means 50\% GPU contention, and ``c6m3600'' means contention with 6 CPU cores and 3600 memory bandwidth. RMSE means root squared mean squared error. We use ``our model/baseline'' formats to show the result.}
  \label{table:tracker_latency_model}
  \scalebox{0.9}{
  \begin{tabular}{|p{0.9in}|p{0.55in}|p{0.6in}|p{0.8in}|}
     \hline
                         & RMSE~(no) & RMSE~(g50) & RMSE~(c6m3600)\\
     \hline
     Medianflow\_ds1     & 11.98/17.86  & 6.37/19.83    & 14.81/44.41   \\
     Medianflow\_ds2     & 6.80/12.37   & 3.14/12.90    & 9.91/26.96   \\
     Medianflow\_ds4     & 7.32/12.30   & 3.72/12.43    & 11.14/26.18    \\
     KCF\_ds4            & 32.23/41.24  & 23.81/43.38   & 41.46/81.37       \\
     CSRT\_ds4           & 46.66/92.54  & 44.17/102.98  & 78.97/179.77      \\
     Dense\_ds4          & 14.15/24.45  & 6.45/26.32    & 11.75/53.26       \\
     \hline
  \end{tabular}
  }
\end{table}

\subsubsection{Tracker latency prediction model:}

We set up the baseline approach, which always predicts a constant latency for each tracker using the averaged latency across all frames under specific contention levels.

As seen from Table~\ref{table:tracker_latency_model}, we can largely reduce the prediction root mean squared error (RMSE) on the test dataset. Some trackers return high prediction RMSE (CSRT for example). The reason is some trackers may have unstable latency under high contention levels. We leave further exploration of this as our future work.

\subsection{Overhead: Switching, Scheduler and Online Components}

\begin{figure}[t]
    \centering
    \includegraphics[width=0.82\linewidth]{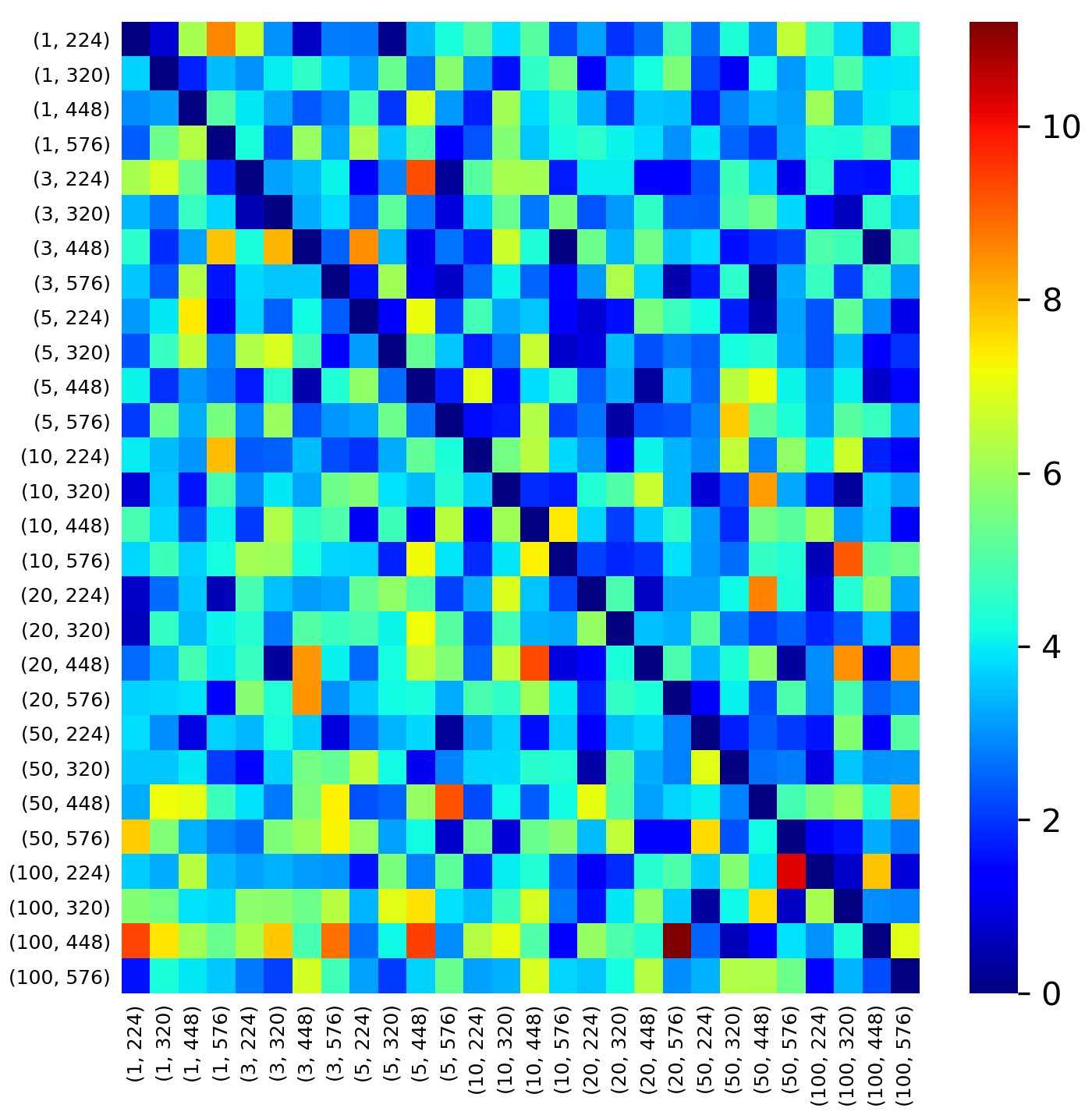}
    \vspace{-0.5em}
    \caption{Heatmap view of switching latency among AB with varying ``numbers of proposals'' and input ``shapes''.}
    \label{fig:switchingheatmap_DNN} 
    \vspace{-1em}
\end{figure}

Figure~\ref{fig:switchingheatmap_DNN} shows a visualization of a heatmap of the switching latency from any approximation branch of the detection framework to another, especially inside the detection DNN. This is an average of 10 such switches and the switching latency is defined as the difference of the execution time of the first frame in the new branches over that of following frames.
We can find that switching latency is bounded by 12 milliseconds.
\begin{table}[th]
  \centering
  \caption{Latency percentage of different parts in \name system. These are measured with zero contention.}
  \label{table:overhead_percentage}
  \scalebox{0.9}{
  \begin{tabular}{|p{0.56in}|p{0.7in}|p{0.55in}|p{0.55in}|} 
     \hline
     User requirement & Scheduler overhead~(\%)  & Detection latency~(\%) & Tracking latency~(\%) \\
     \hline
     80ms     & 0.82\% & 59.65\%  & 39.53\% \\ 
     100ms     & 0.62\%   & 66.63\%   & 32.75\% \\ 
     150ms     & 0.52\%   & 69.10\%   & 30.38\% \\ 
     \hline
  \end{tabular}
  }
\end{table}

The scheduler overhead comes from the accuracy and latency prediction models, as well as the contention sensor. Generally, the overhead of the scheduler is 11.09 milliseconds per execution. Since we will only execute the scheduler one time during all $si$ frames, the average overhead of the scheduler is under 1 millisecond in most cases. This is supported by the profiling results in Table~\ref{table:overhead_percentage}. The overhead of the scheduler occupies less than 1\% of the total latency, suggesting that our scheduler is sufficiently fast to satisfy the requirement of an online system.
\section{Discussion and Future Work}

\noindent\textbf{Relation to the OS and Standing on the Contention.} \name is positioned as a user-level application and does not change the OS's orchestration. Thus, we design \name to treat contention in a black-box manner. The advantage of such standing makes \name easier to implement, free from the OS restrictions, and reduce the complexity of contention scenarios by observing the impact of contention on \name's latency. However due to the black-box method, \name has the limitation of not knowing the exact contention details to adapt accordingly. Different contention scenarios may have the same impact on \name's latency and however \name will not be able to differentiate them, leading to potential sub-optimal adaptation. Future work may add OS privilege, observe the true contention levels from the OS, and even control the contention as well. The marginal benefit over the cost and lost features is yet to be revealed.

\noindent\textbf{Contention Scenarios.} We evaluate \name mainly with GPU contention from the synthetic CG because the object detector mainly runs on the GPU and the detection latency is much higher than the tracker latency. \name has the limitation on the contention source and scenarios, though we also include a case study with a real App. Future work may evaluate with various mobile background workloads from the real trace data and study the propagation effect of \name to the OS or other Apps.

\noindent\textbf{Generalize to Other Detection Methods.} Our adaptive detection framework allows any architectures of detection DNNs beyond FRCNN as long as we can expose the tuning knobs from them. The choice of the detection method can be another approximate knob as well.

\noindent\textbf{Features of Performance Prediction Models.} In \name, we consider the content features motivated by Figure~\ref{fig:tracker_latency_nobjs_feature}, \ref{fig:tracker_latency_sizes_feature} and \ref{fig:movement_ablation} with clear individual impact. We have considered the low-level content feature like the edges in the video frame but it does not have clear impact and thus has been excluded. More features can be introduced to improve the accuracy and latency prediction models, like object type, shape deformation, context, etc.

\noindent\textbf{Reinforcement Learning (RL) for the Scheduler.} RL based models can be an alternative method for implementing the scheduler. RL models learn to make schedule decisions based on training data and can potentially outperform the rule based policies.

\noindent\textbf{Using Neural Networks to Infer Configurations.} We leverage the sampling techniques to reduce the searching cost of configurations. Neural networks can be used to improve the searching efficiency if we disable the sampling and allow more flexible choices.

\noindent\textbf{Evaluation Dataset and Devices.}
We plan to further evaluate \name on data-sets with high resolution videos (\eg 1080p and 2K videos) and other mobile platforms, \textit{e.g.}, devices of lower computation power and smartphones.

\section{Related Work}
\label{sec_related_work}

\vspace{0.2em}
\noindent \textbf{Object Detection}.
Object detection is a well established topic in computer vision and has made recent progress, thanks to DNNs. DNN-based object detectors can be summarized into two categories: single-stage detectors, such as YOLO series~\cite{redmon2016you,redmon2018yolov3}, SSD~\cite{liu2016ssd}, RetinaNet~\cite{lin2017focal}, and two-stage detectors, such as Faster-RCNN~\cite{ren2015faster}, R-FCN~\cite{dai2016r}, Cascade R-CNN~\cite{shuai2018cascaded}. Single-stage detectors classify a dense set of grids, while two-stage detectors focus on a sparse set of region proposals oftentimes producing a separate region proposal network (RPN). Two-stage detectors are usually more accurate with reduced efficiency. While these detectors operate on single images, several recent works seek to extend them to the task of video object detection~\cite{kang2017t,zhu2017flow,feichtenhofer2017detect,zhu2018towards}. A key idea behind these methods is to explore the temporal continuity of videos, where motion tracking is used to enhance the video object detection performance. We used this similar idea for our system, \name.

\vspace{0.2em}
\noindent \textbf{Efficient DNNs for Mobile Applications}
There is an emerging interest to develop efficient DNNs for mobile vision applications. An important line of research is to identify lightweight network architectures with low computational cost. Examples include manually designed MobileNet family ~\cite{howard2017mobilenets,sandler2018mobilenetv2}, SqueezeNet~\cite{iandola2016squeezenet}, and ShuffleNet~\cite{zhang2018shufflenet}, as well as the more recent MNasNet~\cite{tan2019mnasnet}, MobileNetV3~\cite{howard2019searching}, FBNet~\cite{wu2019fbnet}, and EfficientNet~\cite{tan2019efficientnet} produced by neural architecture search~\cite{nas}. Other techniques include the removal of redundant parameters (network pruning)~\cite{han2015deep,li2017pruning,luo2018thinet,hu2016network}, the quantization of parameters (network quantization)~\cite{hubara2016binarized, rastegari2016xnor,hubara2017quantized}. While these architectures and techniques are primarily designed for image classification, they can be used as a building block for efficient object detectors~\cite{wang2018pelee,tan2019efficientdet}. \textit{In spite of the efficiency, none of these approaches can adapt to contention and content during runtime.} 

\vspace{0.2em}
\noindent \textbf{Adaptive Inference for DNNs}
The early work on anytime prediction~\cite{zilberstein1996using} presents the first set of methods that are adaptive to a computing budget at runtime. This idea was recently revisited in the computer vision community using DNNs. For example, at inference time, a DNN can choose to drop certain operations~\cite{wu2018blockdrop,veit2018convolutional}, or to select one of the many exits~\cite{huang2017multi,yang2020resolution} or branches~\cite{teerapittayanon2016branchynet}, based on the image content or a given latency budget. Unfortunately, none of these approaches is designed for object detection. Besides, they do not consider the modeling of resource contention. In parallel to these developments, several recent work in the systems community also seek to build adaptive inference systems for DNNs. For example, NestDNN~\cite{fang2018nestdnn} uses network pruning to convert a static DNN into multiple DNNs, and dynamically selects from these to fit the resource requirement for image classification. AdaScale~\cite{chin2019adascale} learns to adaptively change the input shape of an object detection DNN, in order to achieve a latency-accuracy tradeoff. In comparison to NestDNN, our system \name uses a single adaptive DNN model for object detection. In contrast to AdaScale, \name considers a joint adaptation of video content and resource contention.

\vspace{0.2em}
\noindent \textbf{Approximate Video Analytics}. Our work shares similar ideas to several recent works in video analytics in the systems community. For example, server-side solutions like VideoStorm~\cite{videostorm}, Chameleon~\cite{jiang2018chameleon}, and Focus~\cite{hsieh2018focus} exploit various configurations and DNN models to optimize video analytics queries, but they also require loading of multiple models at the same time, which are challenging in resource-constrained mobile devices. If memory constraints prevent multiple models being co-resident on a device, it is conceivable to send them over the network on demand, using efficient wireless reprogramming~\cite{panta2011efficient}, but the times involved are such that the prediction of which model will be required will have to be done far in advance. Liu \et \cite{liu2019edge} explore the offloading of object detection to an edge device in combination with fast on-device tracking for mobile AR. ExCamera~\cite{excamera-nsdi-2017} enables low-latency video processing on the cloud using serverless architecture. 
VideoChef~\cite{xu2018videochef} uses approximation knobs of traditional video preprocessing filters in a content-aware manner. It cannot handle object detection and is not applicable to DNN-based video processing. 
A recent work in this space called MARLIN~\cite{apicharttrisorn2019frugal} shows that for AR applications, instead of continuously running the detection DNN, they can decide when to run their specially designed lightweight trackers. We take the idea of running the tracker at a configurable interval $si$, but we use traditional trackers. 

\section{Conclusion}
In this paper, we present \name, a single model adaptive system for video object detection in a video content-aware and resource contention-aware fashion, focusing on resource-constrained mobile/embedded devices. The joint context and content aware mechanisms make \name can adaptively respond to both content and context changes while most baseline models cannot respond to any changes or only one change. Case studies have shown our \name can adjust to different scenarios with best accuracy and least latency over previous models. Further, We contrast \name with multiple baselines, including AdaScale, a content-aware adaptive server-based video object detection system, and YOLOv3, a single-stage objection system known for its efficiency on the ImageNet VID dataset. We find that \name is 52.9\% lower in latency and 11.1\% higher in the accuracy metric over YOLOv3 and outperforms all baselines with significantly lower switching overhead, stemming from its single model design. Results on ImageNet VID dataset proves the efficiency and effectiveness of our \name, highlighting \name's promise in enabling latency-sensitive applications, pushing the frontiers of ever-evolving AR/MR (mixed reality) experiences instantiated on embedded platforms.

\section*{ACKNOWLEDGMENTS}
We thank the anonymous reviewers and shepherd for their valuable comments to improve the quality of this paper. This material is based in part upon work supported by the National Science Foundation under Grant Number CNS-1527262, Army Research Lab under Contract number W911NF-20-2-0026, the Lilly Endowment (Wabash Heartland Innovation Network, WHIN-Purdue), and gift funding from Adobe Research. Yin Li acknowledges the support by the UW VCRGE with funding from WARF. Any opinions, findings, and conclusions or recommendations expressed in this material are those of the authors and do not necessarily reflect the views of the sponsors.

\bibliographystyle{ACM-Reference-Format}
\bibliography{main}


\begin{thebibliography}{75}


\ifx \showCODEN    \undefined \def \showCODEN     #1{\unskip}     \fi
\ifx \showDOI      \undefined \def \showDOI       #1{#1}\fi
\ifx \showISBNx    \undefined \def \showISBNx     #1{\unskip}     \fi
\ifx \showISBNxiii \undefined \def \showISBNxiii  #1{\unskip}     \fi
\ifx \showISSN     \undefined \def \showISSN      #1{\unskip}     \fi
\ifx \showLCCN     \undefined \def \showLCCN      #1{\unskip}     \fi
\ifx \shownote     \undefined \def \shownote      #1{#1}          \fi
\ifx \showarticletitle \undefined \def \showarticletitle #1{#1}   \fi
\ifx \showURL      \undefined \def \showURL       {\relax}        \fi
\providecommand\bibfield[2]{#2}
\providecommand\bibinfo[2]{#2}
\providecommand\natexlab[1]{#1}
\providecommand\showeprint[2][]{arXiv:#2}

\bibitem[\protect\citeauthoryear{Ansel, Wong, Chan, Olszewski, Edelman, and
  Amarasinghe}{Ansel et~al\mbox{.}}{2011}]%
        {ansel2011language}
\bibfield{author}{\bibinfo{person}{Jason Ansel}, \bibinfo{person}{Yee~Lok
  Wong}, \bibinfo{person}{Cy Chan}, \bibinfo{person}{Marek Olszewski},
  \bibinfo{person}{Alan Edelman}, {and} \bibinfo{person}{Saman Amarasinghe}.}
  \bibinfo{year}{2011}\natexlab{}.
\newblock \showarticletitle{Language and compiler support for auto-tuning
  variable-accuracy algorithms}. In \bibinfo{booktitle}{\emph{International
  Symposium on Code Generation and Optimization (CGO 2011)}}. IEEE,
  \bibinfo{pages}{85--96}.
\newblock


\bibitem[\protect\citeauthoryear{Apicharttrisorn, Ran, Chen, Krishnamurthy, and
  Roy-Chowdhury}{Apicharttrisorn et~al\mbox{.}}{2019}]%
        {apicharttrisorn2019frugal}
\bibfield{author}{\bibinfo{person}{Kittipat Apicharttrisorn},
  \bibinfo{person}{Xukan Ran}, \bibinfo{person}{Jiasi Chen},
  \bibinfo{person}{Srikanth~V Krishnamurthy}, {and} \bibinfo{person}{Amit~K
  Roy-Chowdhury}.} \bibinfo{year}{2019}\natexlab{}.
\newblock \showarticletitle{Frugal following: Power thrifty object detection
  and tracking for mobile augmented reality}. In
  \bibinfo{booktitle}{\emph{Proceedings of the Conference on Embedded Networked
  Sensor Systems (SenSys)}}. \bibinfo{pages}{96--109}.
\newblock


\bibitem[\protect\citeauthoryear{Bulat}{Bulat}{2020}]%
        {background_ios_react}
\bibfield{author}{\bibinfo{person}{Ross Bulat}.}
  \bibinfo{year}{2020}\natexlab{}.
\newblock \bibinfo{title}{{React Native: Background Task Management in iOS}}.
\newblock
\newblock
\urldef\tempurl%
\url{https://medium.com/@rossbulat/react-native-background-task-management-in-ios-d0f05ae53cc5}
\showURL{%
\tempurl}


\bibitem[\protect\citeauthoryear{Che, Boyer, Meng, Tarjan, Sheaffer, Lee, and
  Skadron}{Che et~al\mbox{.}}{2009}]%
        {che2009rodinia}
\bibfield{author}{\bibinfo{person}{Shuai Che}, \bibinfo{person}{Michael Boyer},
  \bibinfo{person}{Jiayuan Meng}, \bibinfo{person}{David Tarjan},
  \bibinfo{person}{Jeremy~W Sheaffer}, \bibinfo{person}{Sang-Ha Lee}, {and}
  \bibinfo{person}{Kevin Skadron}.} \bibinfo{year}{2009}\natexlab{}.
\newblock \showarticletitle{Rodinia: A benchmark suite for heterogeneous
  computing}. In \bibinfo{booktitle}{\emph{2009 IEEE international symposium on
  workload characterization (IISWC)}}. Ieee, \bibinfo{pages}{44--54}.
\newblock


\bibitem[\protect\citeauthoryear{Chen, Ravindranath, Deng, Bahl, and
  Balakrishnan}{Chen et~al\mbox{.}}{2015}]%
        {chen2015glimpse}
\bibfield{author}{\bibinfo{person}{Tiffany Yu-Han Chen}, \bibinfo{person}{Lenin
  Ravindranath}, \bibinfo{person}{Shuo Deng}, \bibinfo{person}{Paramvir Bahl},
  {and} \bibinfo{person}{Hari Balakrishnan}.} \bibinfo{year}{2015}\natexlab{}.
\newblock \showarticletitle{Glimpse: Continuous, real-time object recognition
  on mobile devices}. In \bibinfo{booktitle}{\emph{Proceedings of the ACM
  Conference on Embedded Networked Sensor Systems (SenSys)}}.
  \bibinfo{pages}{155--168}.
\newblock


\bibitem[\protect\citeauthoryear{Chin, Ding, and Marculescu}{Chin
  et~al\mbox{.}}{2019}]%
        {chin2019adascale}
\bibfield{author}{\bibinfo{person}{Ting-Wu Chin}, \bibinfo{person}{Ruizhou
  Ding}, {and} \bibinfo{person}{Diana Marculescu}.}
  \bibinfo{year}{2019}\natexlab{}.
\newblock \showarticletitle{{AdaScale}: Towards real-time video object
  detection using adaptive scaling}. In \bibinfo{booktitle}{\emph{Proceedings
  of the Conference on Machine Learning and Systems (SysML)}}.
\newblock


\bibitem[\protect\citeauthoryear{Company}{Company}{2020}]%
        {pokemongo}
\bibfield{author}{\bibinfo{person}{The~Pokemon Company}.}
  \bibinfo{year}{2020}\natexlab{}.
\newblock \bibinfo{title}{{Pokémon GO | Augmented Reality Mobile Game}}.
\newblock
\newblock
\urldef\tempurl%
\url{https://pokemongolive.com/en/}
\showURL{%
\tempurl}


\bibitem[\protect\citeauthoryear{Corporation}{Corporation}{2018}]%
        {tx2}
\bibfield{author}{\bibinfo{person}{NVIDIA Corporation}.}
  \bibinfo{year}{2018}\natexlab{}.
\newblock \bibinfo{booktitle}{\emph{Jetson TX2 Module}}.
\newblock
\urldef\tempurl%
\url{https://developer.nvidia.com/embedded/buy/jetson-tx2}
\showURL{%
Retrieved May 5, 2020 from \tempurl}


\bibitem[\protect\citeauthoryear{Dai, Li, He, and Sun}{Dai
  et~al\mbox{.}}{2016}]%
        {dai2016r}
\bibfield{author}{\bibinfo{person}{Jifeng Dai}, \bibinfo{person}{Yi Li},
  \bibinfo{person}{Kaiming He}, {and} \bibinfo{person}{Jian Sun}.}
  \bibinfo{year}{2016}\natexlab{}.
\newblock \showarticletitle{R-{FCN}: Object detection via region-based fully
  convolutional networks}. In \bibinfo{booktitle}{\emph{Proceedings of the
  Advances in Neural Information Processing Systems (NeurIPS)}}.
  \bibinfo{pages}{379--387}.
\newblock


\bibitem[\protect\citeauthoryear{Delimitrou and Kozyrakis}{Delimitrou and
  Kozyrakis}{2013}]%
        {delimitrou2013}
\bibfield{author}{\bibinfo{person}{Christina Delimitrou} {and}
  \bibinfo{person}{Christos Kozyrakis}.} \bibinfo{year}{2013}\natexlab{}.
\newblock \showarticletitle{ibench: Quantifying interference for datacenter
  applications}. In \bibinfo{booktitle}{\emph{2013 IEEE international symposium
  on workload characterization (IISWC)}}. IEEE, \bibinfo{pages}{23--33}.
\newblock


\bibitem[\protect\citeauthoryear{Developer}{Developer}{2019a}]%
        {background_android}
\bibfield{author}{\bibinfo{person}{Android Developer}.}
  \bibinfo{year}{2019}\natexlab{a}.
\newblock \bibinfo{title}{{Guide to background processing: Android}}.
\newblock
\newblock
\urldef\tempurl%
\url{https://developer.android.com/guide/background}
\showURL{%
\tempurl}


\bibitem[\protect\citeauthoryear{Developer}{Developer}{2019b}]%
        {background_ios}
\bibfield{author}{\bibinfo{person}{Apple Developer}.}
  \bibinfo{year}{2019}\natexlab{b}.
\newblock \bibinfo{title}{{Services provided by an app that require it to run
  in the background}}.
\newblock
\newblock
\urldef\tempurl%
\url{https://developer.apple.com/documentation/bundleresources/information_property_list/uibackgroundmodes}
\showURL{%
\tempurl}


\bibitem[\protect\citeauthoryear{Ding, Ansel, Veeramachaneni, Shen, O’Reilly,
  and Amarasinghe}{Ding et~al\mbox{.}}{2015}]%
        {ding2015autotuning}
\bibfield{author}{\bibinfo{person}{Yufei Ding}, \bibinfo{person}{Jason Ansel},
  \bibinfo{person}{Kalyan Veeramachaneni}, \bibinfo{person}{Xipeng Shen},
  \bibinfo{person}{Una-May O’Reilly}, {and} \bibinfo{person}{Saman
  Amarasinghe}.} \bibinfo{year}{2015}\natexlab{}.
\newblock \showarticletitle{Autotuning algorithmic choice for input
  sensitivity}.
\newblock \bibinfo{journal}{\emph{ACM SIGPLAN Notices}} \bibinfo{volume}{50},
  \bibinfo{number}{6} (\bibinfo{year}{2015}), \bibinfo{pages}{379--390}.
\newblock


\bibitem[\protect\citeauthoryear{Elsken, Metzen, and Hutter}{Elsken
  et~al\mbox{.}}{2019}]%
        {nas}
\bibfield{author}{\bibinfo{person}{Thomas Elsken}, \bibinfo{person}{Jan~Hendrik
  Metzen}, {and} \bibinfo{person}{Frank Hutter}.}
  \bibinfo{year}{2019}\natexlab{}.
\newblock \showarticletitle{Neural Architecture Search: A Survey}.
\newblock \bibinfo{journal}{\emph{Journal of Machine Learning Research}}
  \bibinfo{volume}{20}, \bibinfo{number}{55} (\bibinfo{year}{2019}),
  \bibinfo{pages}{1--21}.
\newblock


\bibitem[\protect\citeauthoryear{Fang, Zeng, and Zhang}{Fang
  et~al\mbox{.}}{2018}]%
        {fang2018nestdnn}
\bibfield{author}{\bibinfo{person}{Biyi Fang}, \bibinfo{person}{Xiao Zeng},
  {and} \bibinfo{person}{Mi Zhang}.} \bibinfo{year}{2018}\natexlab{}.
\newblock \showarticletitle{{NestDNN}: Resource-aware multi-tenant on-device
  deep learning for continuous mobile vision}. In
  \bibinfo{booktitle}{\emph{Proceedings of the Annual International Conference
  on Mobile Computing and Networking (MobiCom)}}. \bibinfo{pages}{115--127}.
\newblock


\bibitem[\protect\citeauthoryear{Farneb{\"a}ck}{Farneb{\"a}ck}{2003}]%
        {farneback2003two}
\bibfield{author}{\bibinfo{person}{Gunnar Farneb{\"a}ck}.}
  \bibinfo{year}{2003}\natexlab{}.
\newblock \showarticletitle{Two-frame motion estimation based on polynomial
  expansion}. In \bibinfo{booktitle}{\emph{Proceedings of Scandinavian
  Conference on Image Analysis}}. \bibinfo{pages}{363--370}.
\newblock


\bibitem[\protect\citeauthoryear{Feichtenhofer, Pinz, and
  Zisserman}{Feichtenhofer et~al\mbox{.}}{2017}]%
        {feichtenhofer2017detect}
\bibfield{author}{\bibinfo{person}{Christoph Feichtenhofer},
  \bibinfo{person}{Axel Pinz}, {and} \bibinfo{person}{Andrew Zisserman}.}
  \bibinfo{year}{2017}\natexlab{}.
\newblock \showarticletitle{Detect to track and track to detect}. In
  \bibinfo{booktitle}{\emph{Proceedings of the IEEE International Conference on
  Computer Vision (ICCV)}}. \bibinfo{pages}{3038--3046}.
\newblock


\bibitem[\protect\citeauthoryear{Fouladi, Wahby, Shacklett, Balasubramaniam,
  Zeng, Bhalerao, Sivaraman, Porter, and Winstein}{Fouladi
  et~al\mbox{.}}{2017}]%
        {excamera-nsdi-2017}
\bibfield{author}{\bibinfo{person}{Sadjad Fouladi}, \bibinfo{person}{Riad~S
  Wahby}, \bibinfo{person}{Brennan Shacklett}, \bibinfo{person}{Karthikeyan
  Balasubramaniam}, \bibinfo{person}{William Zeng}, \bibinfo{person}{Rahul
  Bhalerao}, \bibinfo{person}{Anirudh Sivaraman}, \bibinfo{person}{George
  Porter}, {and} \bibinfo{person}{Keith Winstein}.}
  \bibinfo{year}{2017}\natexlab{}.
\newblock \showarticletitle{Encoding, Fast and Slow: Low-Latency Video
  Processing Using Thousands of Tiny Threads.}. In
  \bibinfo{booktitle}{\emph{Proceedings of the Symposium on Networked Systems
  Design and Implementation (NSDI)}}. \bibinfo{pages}{363--376}.
\newblock


\bibitem[\protect\citeauthoryear{Guha, Mishra, Roy, and Schrijvers}{Guha
  et~al\mbox{.}}{2016}]%
        {guha2016robust}
\bibfield{author}{\bibinfo{person}{Sudipto Guha}, \bibinfo{person}{Nina
  Mishra}, \bibinfo{person}{Gourav Roy}, {and} \bibinfo{person}{Okke
  Schrijvers}.} \bibinfo{year}{2016}\natexlab{}.
\newblock \showarticletitle{Robust random cut forest based anomaly detection on
  streams}. In \bibinfo{booktitle}{\emph{Proceedings of the International
  Conference on Machine Learning (ICML)}}. \bibinfo{pages}{2712--2721}.
\newblock


\bibitem[\protect\citeauthoryear{Han, Mao, and Dally}{Han
  et~al\mbox{.}}{2016a}]%
        {han2015deep}
\bibfield{author}{\bibinfo{person}{Song Han}, \bibinfo{person}{Huizi Mao},
  {and} \bibinfo{person}{William~J Dally}.} \bibinfo{year}{2016}\natexlab{a}.
\newblock \showarticletitle{Deep compression: Compressing deep neural networks
  with pruning, trained quantization and huffman coding}.
\newblock  (\bibinfo{year}{2016}), \bibinfo{pages}{1--13}.
\newblock


\bibitem[\protect\citeauthoryear{Han, Shen, Philipose, Agarwal, Wolman, and
  Krishnamurthy}{Han et~al\mbox{.}}{2016b}]%
        {mcdnn}
\bibfield{author}{\bibinfo{person}{Seungyeop Han}, \bibinfo{person}{Haichen
  Shen}, \bibinfo{person}{Matthai Philipose}, \bibinfo{person}{Sharad Agarwal},
  \bibinfo{person}{Alec Wolman}, {and} \bibinfo{person}{Arvind Krishnamurthy}.}
  \bibinfo{year}{2016}\natexlab{b}.
\newblock \showarticletitle{{MCDNN}: An approximation-based execution framework
  for deep stream processing under resource constraints}. In
  \bibinfo{booktitle}{\emph{Proceedings of the Annual International Conference
  on Mobile Systems, Applications, and Services (MobiSys)}}.
  \bibinfo{pages}{123--136}.
\newblock


\bibitem[\protect\citeauthoryear{Henriques, Caseiro, Martins, and
  Batista}{Henriques et~al\mbox{.}}{2014}]%
        {KCF}
\bibfield{author}{\bibinfo{person}{J.~F. Henriques}, \bibinfo{person}{R.
  Caseiro}, \bibinfo{person}{P. Martins}, {and} \bibinfo{person}{J. Batista}.}
  \bibinfo{year}{2014}\natexlab{}.
\newblock \showarticletitle{High-Speed Tracking with Kernelized Correlation
  Filters}.
\newblock \bibinfo{journal}{\emph{IEEE Transactions on Pattern Analysis and
  Machine Intelligence}} \bibinfo{volume}{37}, \bibinfo{number}{3}
  (\bibinfo{year}{2014}), \bibinfo{pages}{583--596}.
\newblock


\bibitem[\protect\citeauthoryear{Howard, Sandler, Chu, Chen, Chen, Tan, Wang,
  Zhu, Pang, Vasudevan, et~al\mbox{.}}{Howard et~al\mbox{.}}{2019}]%
        {howard2019searching}
\bibfield{author}{\bibinfo{person}{Andrew Howard}, \bibinfo{person}{Mark
  Sandler}, \bibinfo{person}{Grace Chu}, \bibinfo{person}{Liang-Chieh Chen},
  \bibinfo{person}{Bo Chen}, \bibinfo{person}{Mingxing Tan},
  \bibinfo{person}{Weijun Wang}, \bibinfo{person}{Yukun Zhu},
  \bibinfo{person}{Ruoming Pang}, \bibinfo{person}{Vijay Vasudevan},
  {et~al\mbox{.}}} \bibinfo{year}{2019}\natexlab{}.
\newblock \showarticletitle{Searching for {MobileNetV3}}. In
  \bibinfo{booktitle}{\emph{Proceedings of the IEEE International Conference on
  Computer Vision (ICCV)}}. \bibinfo{pages}{1314--1324}.
\newblock


\bibitem[\protect\citeauthoryear{Howard, Zhu, Chen, Kalenichenko, Wang, Weyand,
  Andreetto, and Adam}{Howard et~al\mbox{.}}{2017}]%
        {howard2017mobilenets}
\bibfield{author}{\bibinfo{person}{Andrew~G Howard}, \bibinfo{person}{Menglong
  Zhu}, \bibinfo{person}{Bo Chen}, \bibinfo{person}{Dmitry Kalenichenko},
  \bibinfo{person}{Weijun Wang}, \bibinfo{person}{Tobias Weyand},
  \bibinfo{person}{Marco Andreetto}, {and} \bibinfo{person}{Hartwig Adam}.}
  \bibinfo{year}{2017}\natexlab{}.
\newblock \showarticletitle{{MobileNets}: Efficient convolutional neural
  networks for mobile vision applications}.
\newblock \bibinfo{journal}{\emph{arXiv preprint arXiv:1704.04861}}
  (\bibinfo{year}{2017}).
\newblock


\bibitem[\protect\citeauthoryear{Hsieh, Ananthanarayanan, Bodik, Venkataraman,
  Bahl, Philipose, Gibbons, and Mutlu}{Hsieh et~al\mbox{.}}{2018}]%
        {hsieh2018focus}
\bibfield{author}{\bibinfo{person}{Kevin Hsieh}, \bibinfo{person}{Ganesh
  Ananthanarayanan}, \bibinfo{person}{Peter Bodik}, \bibinfo{person}{Shivaram
  Venkataraman}, \bibinfo{person}{Paramvir Bahl}, \bibinfo{person}{Matthai
  Philipose}, \bibinfo{person}{Phillip~B Gibbons}, {and} \bibinfo{person}{Onur
  Mutlu}.} \bibinfo{year}{2018}\natexlab{}.
\newblock \showarticletitle{Focus: Querying large video datasets with low
  latency and low cost}. In \bibinfo{booktitle}{\emph{13th $\{$USENIX$\}$
  Symposium on Operating Systems Design and Implementation ($\{$OSDI$\}$ 18)}}.
  \bibinfo{pages}{269--286}.
\newblock


\bibitem[\protect\citeauthoryear{Hu, Peng, Tai, and Tang}{Hu
  et~al\mbox{.}}{2016}]%
        {hu2016network}
\bibfield{author}{\bibinfo{person}{Hengyuan Hu}, \bibinfo{person}{Rui Peng},
  \bibinfo{person}{Yu-Wing Tai}, {and} \bibinfo{person}{Chi-Keung Tang}.}
  \bibinfo{year}{2016}\natexlab{}.
\newblock \showarticletitle{Network trimming: A data-driven neuron pruning
  approach towards efficient deep architectures}.
\newblock \bibinfo{journal}{\emph{arXiv preprint arXiv:1607.03250}}
  (\bibinfo{year}{2016}).
\newblock


\bibitem[\protect\citeauthoryear{Huang, Chen, Li, Wu, van~der Maaten, and
  Weinberger}{Huang et~al\mbox{.}}{2018}]%
        {huang2017multi}
\bibfield{author}{\bibinfo{person}{Gao Huang}, \bibinfo{person}{Danlu Chen},
  \bibinfo{person}{Tianhong Li}, \bibinfo{person}{Felix Wu},
  \bibinfo{person}{Laurens van~der Maaten}, {and} \bibinfo{person}{Kilian~Q
  Weinberger}.} \bibinfo{year}{2018}\natexlab{}.
\newblock \showarticletitle{Multi-scale dense networks for resource efficient
  image classification}. In \bibinfo{booktitle}{\emph{Proceedings of
  International Conference on Learning Representations (ICLR)}}.
\newblock


\bibitem[\protect\citeauthoryear{Hubara, Courbariaux, Soudry, El-Yaniv, and
  Bengio}{Hubara et~al\mbox{.}}{2016}]%
        {hubara2016binarized}
\bibfield{author}{\bibinfo{person}{Itay Hubara}, \bibinfo{person}{Matthieu
  Courbariaux}, \bibinfo{person}{Daniel Soudry}, \bibinfo{person}{Ran
  El-Yaniv}, {and} \bibinfo{person}{Yoshua Bengio}.}
  \bibinfo{year}{2016}\natexlab{}.
\newblock \showarticletitle{Binarized neural networks}. In
  \bibinfo{booktitle}{\emph{Proceedings of the Advances in Neural Information
  Processing Systems (NeurIPS)}}. \bibinfo{pages}{4107--4115}.
\newblock


\bibitem[\protect\citeauthoryear{Hubara, Courbariaux, Soudry, El-Yaniv, and
  Bengio}{Hubara et~al\mbox{.}}{2017}]%
        {hubara2017quantized}
\bibfield{author}{\bibinfo{person}{Itay Hubara}, \bibinfo{person}{Matthieu
  Courbariaux}, \bibinfo{person}{Daniel Soudry}, \bibinfo{person}{Ran
  El-Yaniv}, {and} \bibinfo{person}{Yoshua Bengio}.}
  \bibinfo{year}{2017}\natexlab{}.
\newblock \showarticletitle{Quantized Neural Networks: Training Neural Networks
  with Low Precision Weights and Activations}.
\newblock \bibinfo{journal}{\emph{Journal of Machine Learning Research}}
  \bibinfo{volume}{18} (\bibinfo{year}{2017}), \bibinfo{pages}{187--1}.
\newblock


\bibitem[\protect\citeauthoryear{Iandola, Han, Moskewicz, Ashraf, Dally, and
  Keutzer}{Iandola et~al\mbox{.}}{2016}]%
        {iandola2016squeezenet}
\bibfield{author}{\bibinfo{person}{Forrest~N Iandola}, \bibinfo{person}{Song
  Han}, \bibinfo{person}{Matthew~W Moskewicz}, \bibinfo{person}{Khalid Ashraf},
  \bibinfo{person}{William~J Dally}, {and} \bibinfo{person}{Kurt Keutzer}.}
  \bibinfo{year}{2016}\natexlab{}.
\newblock \showarticletitle{{SqueezeNet}: {AlexNet-level} accuracy with 50x
  fewer parameters and< 0.5 {MB} model size}. In
  \bibinfo{booktitle}{\emph{Proceedings of International Conference on Learning
  Representations (ICLR)}}. \bibinfo{pages}{1--13}.
\newblock


\bibitem[\protect\citeauthoryear{Jiang, Wong, Canel, Tang, Misra, Kaminsky,
  Kozuch, Pillai, Andersen, and Ganger}{Jiang et~al\mbox{.}}{2018b}]%
        {jiang2018mainstream}
\bibfield{author}{\bibinfo{person}{Angela~H Jiang}, \bibinfo{person}{Daniel L-K
  Wong}, \bibinfo{person}{Christopher Canel}, \bibinfo{person}{Lilia Tang},
  \bibinfo{person}{Ishan Misra}, \bibinfo{person}{Michael Kaminsky},
  \bibinfo{person}{Michael~A Kozuch}, \bibinfo{person}{Padmanabhan Pillai},
  \bibinfo{person}{David~G Andersen}, {and} \bibinfo{person}{Gregory~R
  Ganger}.} \bibinfo{year}{2018}\natexlab{b}.
\newblock \showarticletitle{Mainstream: Dynamic Stem-Sharing for Multi-Tenant
  Video Processing}. In \bibinfo{booktitle}{\emph{Proceedings of USENIX Annual
  Technical Conference (USENIX ATC)}}. \bibinfo{pages}{29--42}.
\newblock


\bibitem[\protect\citeauthoryear{Jiang, Ananthanarayanan, Bodik, Sen, and
  Stoica}{Jiang et~al\mbox{.}}{2018a}]%
        {jiang2018chameleon}
\bibfield{author}{\bibinfo{person}{Junchen Jiang}, \bibinfo{person}{Ganesh
  Ananthanarayanan}, \bibinfo{person}{Peter Bodik}, \bibinfo{person}{Siddhartha
  Sen}, {and} \bibinfo{person}{Ion Stoica}.} \bibinfo{year}{2018}\natexlab{a}.
\newblock \showarticletitle{Chameleon: scalable adaptation of video analytics}.
  In \bibinfo{booktitle}{\emph{Proceedings of the Conference of the ACM Special
  Interest Group on Data Communication (SIGCOMM)}}. \bibinfo{pages}{253--266}.
\newblock


\bibitem[\protect\citeauthoryear{Kalal, Mikolajczyk, and Matas}{Kalal
  et~al\mbox{.}}{2010}]%
        {MedianFlow}
\bibfield{author}{\bibinfo{person}{Zdenek Kalal}, \bibinfo{person}{Krystian
  Mikolajczyk}, {and} \bibinfo{person}{Jiri Matas}.}
  \bibinfo{year}{2010}\natexlab{}.
\newblock \showarticletitle{{Forward-Backward error}: Automatic detection of
  tracking failures}. In \bibinfo{booktitle}{\emph{Proceedings of IEEE
  International Conference on Pattern Recognition (CVPR)}}.
  \bibinfo{pages}{2756--2759}.
\newblock


\bibitem[\protect\citeauthoryear{Kang, Li, Yan, Zeng, Yang, Xiao, Zhang, Wang,
  Wang, Wang, et~al\mbox{.}}{Kang et~al\mbox{.}}{2017}]%
        {kang2017t}
\bibfield{author}{\bibinfo{person}{Kai Kang}, \bibinfo{person}{Hongsheng Li},
  \bibinfo{person}{Junjie Yan}, \bibinfo{person}{Xingyu Zeng},
  \bibinfo{person}{Bin Yang}, \bibinfo{person}{Tong Xiao},
  \bibinfo{person}{Cong Zhang}, \bibinfo{person}{Zhe Wang},
  \bibinfo{person}{Ruohui Wang}, \bibinfo{person}{Xiaogang Wang},
  {et~al\mbox{.}}} \bibinfo{year}{2017}\natexlab{}.
\newblock \showarticletitle{T-{CNN}: Tubelets with convolutional neural
  networks for object detection from videos}.
\newblock \bibinfo{journal}{\emph{IEEE Transactions on Circuits and Systems for
  Video Technology}} \bibinfo{volume}{28}, \bibinfo{number}{10}
  (\bibinfo{year}{2017}), \bibinfo{pages}{2896--2907}.
\newblock


\bibitem[\protect\citeauthoryear{Laurenzano, Hill, Samadi, Mahlke, Mars, and
  Tang}{Laurenzano et~al\mbox{.}}{2016}]%
        {laurenzano2016input}
\bibfield{author}{\bibinfo{person}{Michael~A Laurenzano},
  \bibinfo{person}{Parker Hill}, \bibinfo{person}{Mehrzad Samadi},
  \bibinfo{person}{Scott Mahlke}, \bibinfo{person}{Jason Mars}, {and}
  \bibinfo{person}{Lingjia Tang}.} \bibinfo{year}{2016}\natexlab{}.
\newblock \showarticletitle{Input responsiveness: using canary inputs to
  dynamically steer approximation}. In \bibinfo{booktitle}{\emph{Proceedings of
  the 37th ACM SIGPLAN Conference on Programming Language Design and
  Implementation (PLDI)}}. \bibinfo{pages}{161--176}.
\newblock


\bibitem[\protect\citeauthoryear{Li, Kadav, Durdanovic, Samet, and Graf}{Li
  et~al\mbox{.}}{2017}]%
        {li2017pruning}
\bibfield{author}{\bibinfo{person}{Hao Li}, \bibinfo{person}{Asim Kadav},
  \bibinfo{person}{Igor Durdanovic}, \bibinfo{person}{Hanan Samet}, {and}
  \bibinfo{person}{Hans~Peter Graf}.} \bibinfo{year}{2017}\natexlab{}.
\newblock \showarticletitle{Pruning filters for efficient {ConvNets}}. In
  \bibinfo{booktitle}{\emph{Proceedings of International Conference on Learning
  Representations (ICLR)}}. \bibinfo{pages}{1--13}.
\newblock


\bibitem[\protect\citeauthoryear{Lin, Goyal, Girshick, He, and Doll{\'a}r}{Lin
  et~al\mbox{.}}{2017}]%
        {lin2017focal}
\bibfield{author}{\bibinfo{person}{Tsung-Yi Lin}, \bibinfo{person}{Priya
  Goyal}, \bibinfo{person}{Ross Girshick}, \bibinfo{person}{Kaiming He}, {and}
  \bibinfo{person}{Piotr Doll{\'a}r}.} \bibinfo{year}{2017}\natexlab{}.
\newblock \showarticletitle{Focal loss for dense object detection}. In
  \bibinfo{booktitle}{\emph{Proceedings of the IEEE International Conference on
  Computer Vision (ICCV)}}. \bibinfo{pages}{2980--2988}.
\newblock


\bibitem[\protect\citeauthoryear{Liu, Li, and Gruteser}{Liu
  et~al\mbox{.}}{2019}]%
        {liu2019edge}
\bibfield{author}{\bibinfo{person}{Luyang Liu}, \bibinfo{person}{Hongyu Li},
  {and} \bibinfo{person}{Marco Gruteser}.} \bibinfo{year}{2019}\natexlab{}.
\newblock \showarticletitle{Edge assisted real-time object detection for mobile
  augmented reality}.
\newblock  (\bibinfo{year}{2019}), \bibinfo{pages}{1--16}.
\newblock


\bibitem[\protect\citeauthoryear{Liu, Anguelov, Erhan, Szegedy, Reed, Fu, and
  Berg}{Liu et~al\mbox{.}}{2016}]%
        {liu2016ssd}
\bibfield{author}{\bibinfo{person}{Wei Liu}, \bibinfo{person}{Dragomir
  Anguelov}, \bibinfo{person}{Dumitru Erhan}, \bibinfo{person}{Christian
  Szegedy}, \bibinfo{person}{Scott Reed}, \bibinfo{person}{Cheng-Yang Fu},
  {and} \bibinfo{person}{Alexander~C Berg}.} \bibinfo{year}{2016}\natexlab{}.
\newblock \showarticletitle{{SSD}: Single shot multibox detector}. In
  \bibinfo{booktitle}{\emph{Proceedings of the European conference on Computer
  Vision (ECCV)}}, Vol.~\bibinfo{volume}{9907}. \bibinfo{pages}{21--37}.
\newblock


\bibitem[\protect\citeauthoryear{Luke{\v{z}}i{\v{c}}, Voj{'i}{\v{r}},
  {\v{C}}ehovin~Zajc, Matas, and Kristan}{Luke{\v{z}}i{\v{c}}
  et~al\mbox{.}}{2018}]%
        {CSRT}
\bibfield{author}{\bibinfo{person}{Alan Luke{\v{z}}i{\v{c}}},
  \bibinfo{person}{Tom{'a}{\v{s}} Voj{'i}{\v{r}}}, \bibinfo{person}{Luka
  {\v{C}}ehovin~Zajc}, \bibinfo{person}{Ji{\v{r}}{'i} Matas}, {and}
  \bibinfo{person}{Matej Kristan}.} \bibinfo{year}{2018}\natexlab{}.
\newblock \showarticletitle{Discriminative Correlation Filter Tracker with
  Channel and Spatial Reliability}.
\newblock \bibinfo{journal}{\emph{International Journal of Computer Vision}}
  \bibinfo{volume}{126} (\bibinfo{year}{2018}), \bibinfo{pages}{671--688}.
\newblock


\bibitem[\protect\citeauthoryear{Luo, Zhang, Zhou, Xie, Wu, and Lin}{Luo
  et~al\mbox{.}}{2018}]%
        {luo2018thinet}
\bibfield{author}{\bibinfo{person}{Jian-Hao Luo}, \bibinfo{person}{Hao Zhang},
  \bibinfo{person}{Hong-Yu Zhou}, \bibinfo{person}{Chen-Wei Xie},
  \bibinfo{person}{Jianxin Wu}, {and} \bibinfo{person}{Weiyao Lin}.}
  \bibinfo{year}{2018}\natexlab{}.
\newblock \showarticletitle{{ThiNet}: pruning {CNN} filters for a thinner net}.
\newblock \bibinfo{journal}{\emph{IEEE Transactions on Pattern Analysis and
  Machine Intelligence}} \bibinfo{volume}{41}, \bibinfo{number}{10}
  (\bibinfo{year}{2018}), \bibinfo{pages}{2525--2538}.
\newblock


\bibitem[\protect\citeauthoryear{Maji, Mitra, Zhou, Bagchi, and Verma}{Maji
  et~al\mbox{.}}{2014}]%
        {maji2014}
\bibfield{author}{\bibinfo{person}{Amiya~K Maji}, \bibinfo{person}{Subrata
  Mitra}, \bibinfo{person}{Bowen Zhou}, \bibinfo{person}{Saurabh Bagchi}, {and}
  \bibinfo{person}{Akshat Verma}.} \bibinfo{year}{2014}\natexlab{}.
\newblock \showarticletitle{Mitigating interference in cloud services by
  middleware reconfiguration}. In \bibinfo{booktitle}{\emph{Proceedings of the
  15th International Middleware Conference}}. \bibinfo{pages}{277--288}.
\newblock


\bibitem[\protect\citeauthoryear{Mars, Tang, Hundt, Skadron, and Soffa}{Mars
  et~al\mbox{.}}{2011}]%
        {mars2011bubble}
\bibfield{author}{\bibinfo{person}{Jason Mars}, \bibinfo{person}{Lingjia Tang},
  \bibinfo{person}{Robert Hundt}, \bibinfo{person}{Kevin Skadron}, {and}
  \bibinfo{person}{Mary~Lou Soffa}.} \bibinfo{year}{2011}\natexlab{}.
\newblock \showarticletitle{Bubble-up: Increasing utilization in modern
  warehouse scale computers via sensible co-locations}. In
  \bibinfo{booktitle}{\emph{Proceedings of the 44th annual IEEE/ACM
  International Symposium on Microarchitecture}}. \bibinfo{pages}{248--259}.
\newblock


\bibitem[\protect\citeauthoryear{McCalpin}{McCalpin}{2007}]%
        {McCalpin2007}
\bibfield{author}{\bibinfo{person}{John~D. McCalpin}.}
  \bibinfo{year}{1991-2007}\natexlab{}.
\newblock \bibinfo{booktitle}{\emph{STREAM: Sustainable Memory Bandwidth in
  High Performance Computers}}.
\newblock \bibinfo{type}{{T}echnical {R}eport}.
  \bibinfo{institution}{University of Virginia},
  \bibinfo{address}{Charlottesville, Virginia}.
\newblock
\urldef\tempurl%
\url{http://www.cs.virginia.edu/stream/}
\showURL{%
\tempurl}


\bibitem[\protect\citeauthoryear{McCalpin}{McCalpin}{1995}]%
        {McCalpin1995}
\bibfield{author}{\bibinfo{person}{John~D. McCalpin}.}
  \bibinfo{year}{1995}\natexlab{}.
\newblock \showarticletitle{Memory Bandwidth and Machine Balance in Current
  High Performance Computers}.
\newblock \bibinfo{journal}{\emph{IEEE Computer Society Technical Committee on
  Computer Architecture (TCCA) Newsletter}} (\bibinfo{date}{Dec.}
  \bibinfo{year}{1995}), \bibinfo{pages}{19--25}.
\newblock


\bibitem[\protect\citeauthoryear{Min, Montanari, Mathur, and Kawsar}{Min
  et~al\mbox{.}}{2019}]%
        {min2019closer}
\bibfield{author}{\bibinfo{person}{Chulhong Min}, \bibinfo{person}{Alessandro
  Montanari}, \bibinfo{person}{Akhil Mathur}, {and} \bibinfo{person}{Fahim
  Kawsar}.} \bibinfo{year}{2019}\natexlab{}.
\newblock \showarticletitle{A closer look at quality-aware runtime assessment
  of sensing models in multi-device environments}. In
  \bibinfo{booktitle}{\emph{Proceedings of the 17th Conference on Embedded
  Networked Sensor Systems}}. \bibinfo{pages}{271--284}.
\newblock


\bibitem[\protect\citeauthoryear{Mitra, Gupta, Misailovic, and Bagchi}{Mitra
  et~al\mbox{.}}{2017}]%
        {mitra2017phase}
\bibfield{author}{\bibinfo{person}{Subrata Mitra}, \bibinfo{person}{Manish~K
  Gupta}, \bibinfo{person}{Sasa Misailovic}, {and} \bibinfo{person}{Saurabh
  Bagchi}.} \bibinfo{year}{2017}\natexlab{}.
\newblock \showarticletitle{Phase-aware optimization in approximate computing}.
  In \bibinfo{booktitle}{\emph{2017 IEEE/ACM International Symposium on Code
  Generation and Optimization (CGO)}}. IEEE, \bibinfo{pages}{185--196}.
\newblock


\bibitem[\protect\citeauthoryear{Panta, Bagchi, and Midkiff}{Panta
  et~al\mbox{.}}{2011}]%
        {panta2011efficient}
\bibfield{author}{\bibinfo{person}{Rajesh~Krishna Panta},
  \bibinfo{person}{Saurabh Bagchi}, {and} \bibinfo{person}{Samuel~P Midkiff}.}
  \bibinfo{year}{2011}\natexlab{}.
\newblock \showarticletitle{Efficient incremental code update for sensor
  networks}.
\newblock \bibinfo{journal}{\emph{ACM Transactions on Sensor Networks (TOSN)}}
  \bibinfo{volume}{7}, \bibinfo{number}{4} (\bibinfo{year}{2011}),
  \bibinfo{pages}{1--32}.
\newblock


\bibitem[\protect\citeauthoryear{Rastegari, Ordonez, Redmon, and
  Farhadi}{Rastegari et~al\mbox{.}}{2016}]%
        {rastegari2016xnor}
\bibfield{author}{\bibinfo{person}{Mohammad Rastegari},
  \bibinfo{person}{Vicente Ordonez}, \bibinfo{person}{Joseph Redmon}, {and}
  \bibinfo{person}{Ali Farhadi}.} \bibinfo{year}{2016}\natexlab{}.
\newblock \showarticletitle{{XNOR-Net}: {ImageNet} classification using binary
  convolutional neural networks}. In \bibinfo{booktitle}{\emph{Proceedings of
  the European Conference on Computer Vision (ECCV)}},
  Vol.~\bibinfo{volume}{9908}. \bibinfo{pages}{525--542}.
\newblock


\bibitem[\protect\citeauthoryear{Redmon, Divvala, Girshick, and Farhadi}{Redmon
  et~al\mbox{.}}{2016}]%
        {redmon2016you}
\bibfield{author}{\bibinfo{person}{Joseph Redmon}, \bibinfo{person}{Santosh
  Divvala}, \bibinfo{person}{Ross Girshick}, {and} \bibinfo{person}{Ali
  Farhadi}.} \bibinfo{year}{2016}\natexlab{}.
\newblock \showarticletitle{You only look once: Unified, real-time object
  detection}. In \bibinfo{booktitle}{\emph{Proceedings of the IEEE Conference
  on Computer Vision and Pattern Recognition (CVPR)}}.
  \bibinfo{pages}{779--788}.
\newblock


\bibitem[\protect\citeauthoryear{Redmon and Farhadi}{Redmon and
  Farhadi}{2018}]%
        {redmon2018yolov3}
\bibfield{author}{\bibinfo{person}{Joseph Redmon} {and} \bibinfo{person}{Ali
  Farhadi}.} \bibinfo{year}{2018}\natexlab{}.
\newblock \showarticletitle{{YOLOv3}: An incremental improvement}.
\newblock \bibinfo{journal}{\emph{arXiv preprint arXiv:1804.02767}}
  (\bibinfo{year}{2018}).
\newblock


\bibitem[\protect\citeauthoryear{Ren, He, Girshick, and Sun}{Ren
  et~al\mbox{.}}{2015}]%
        {ren2015faster}
\bibfield{author}{\bibinfo{person}{Shaoqing Ren}, \bibinfo{person}{Kaiming He},
  \bibinfo{person}{Ross Girshick}, {and} \bibinfo{person}{Jian Sun}.}
  \bibinfo{year}{2015}\natexlab{}.
\newblock \showarticletitle{Faster {R}-{CNN}: Towards real-time object
  detection with region proposal networks}. In
  \bibinfo{booktitle}{\emph{Proceedings of the Advances in Neural Information
  Processing Systems (NeurIPS)}}. \bibinfo{pages}{91--99}.
\newblock


\bibitem[\protect\citeauthoryear{Russakovsky, Deng, Su, Krause, Satheesh, Ma,
  Huang, Karpathy, Khosla, Bernstein, Berg, and Fei-Fei}{Russakovsky
  et~al\mbox{.}}{2015}]%
        {ILSVRC15}
\bibfield{author}{\bibinfo{person}{Olga Russakovsky}, \bibinfo{person}{Jia
  Deng}, \bibinfo{person}{Hao Su}, \bibinfo{person}{Jonathan Krause},
  \bibinfo{person}{Sanjeev Satheesh}, \bibinfo{person}{Sean Ma},
  \bibinfo{person}{Zhiheng Huang}, \bibinfo{person}{Andrej Karpathy},
  \bibinfo{person}{Aditya Khosla}, \bibinfo{person}{Michael Bernstein},
  \bibinfo{person}{Alexander~C. Berg}, {and} \bibinfo{person}{Li Fei-Fei}.}
  \bibinfo{year}{2015}\natexlab{}.
\newblock \showarticletitle{{ImageNet} Large Scale Visual Recognition
  Challenge}.
\newblock \bibinfo{journal}{\emph{International Journal of Computer Vision}}
  \bibinfo{volume}{115}, \bibinfo{number}{3} (\bibinfo{year}{2015}),
  \bibinfo{pages}{211--252}.
\newblock


\bibitem[\protect\citeauthoryear{Sandler, Howard, Zhu, Zhmoginov, and
  Chen}{Sandler et~al\mbox{.}}{2018}]%
        {sandler2018mobilenetv2}
\bibfield{author}{\bibinfo{person}{Mark Sandler}, \bibinfo{person}{Andrew
  Howard}, \bibinfo{person}{Menglong Zhu}, \bibinfo{person}{Andrey Zhmoginov},
  {and} \bibinfo{person}{Liang-Chieh Chen}.} \bibinfo{year}{2018}\natexlab{}.
\newblock \showarticletitle{{MobileNetv2}: Inverted residuals and linear
  bottlenecks}. In \bibinfo{booktitle}{\emph{Proceedings of the IEEE Conference
  on Computer Vision and Pattern Recognition (CVPR)}}.
  \bibinfo{pages}{4510--4520}.
\newblock


\bibitem[\protect\citeauthoryear{Shen, Hertzmann, Jia, Paris, Price, Shechtman,
  and Sachs}{Shen et~al\mbox{.}}{2016}]%
        {shen2016automatic}
\bibfield{author}{\bibinfo{person}{Xiaoyong Shen}, \bibinfo{person}{Aaron
  Hertzmann}, \bibinfo{person}{Jiaya Jia}, \bibinfo{person}{Sylvain Paris},
  \bibinfo{person}{Brian Price}, \bibinfo{person}{Eli Shechtman}, {and}
  \bibinfo{person}{Ian Sachs}.} \bibinfo{year}{2016}\natexlab{}.
\newblock \showarticletitle{Automatic portrait segmentation for image
  stylization}. In \bibinfo{booktitle}{\emph{Computer Graphics Forum}},
  Vol.~\bibinfo{volume}{35}. Wiley Online Library, \bibinfo{pages}{93--102}.
\newblock


\bibitem[\protect\citeauthoryear{Shuai, Liu, Zhang, Yang, and Deng}{Shuai
  et~al\mbox{.}}{2018}]%
        {shuai2018cascaded}
\bibfield{author}{\bibinfo{person}{Hui Shuai}, \bibinfo{person}{Qingshan Liu},
  \bibinfo{person}{Kaihua Zhang}, \bibinfo{person}{Jing Yang}, {and}
  \bibinfo{person}{Jiankang Deng}.} \bibinfo{year}{2018}\natexlab{}.
\newblock \showarticletitle{Cascaded Regional Spatio-Temporal Feature-Routing
  Networks for Video Object Detection}.
\newblock \bibinfo{journal}{\emph{IEEE Access}}  \bibinfo{volume}{6}
  (\bibinfo{year}{2018}), \bibinfo{pages}{3096--3106}.
\newblock


\bibitem[\protect\citeauthoryear{Tan, Chen, Pang, Vasudevan, Sandler, Howard,
  and Le}{Tan et~al\mbox{.}}{2019}]%
        {tan2019mnasnet}
\bibfield{author}{\bibinfo{person}{Mingxing Tan}, \bibinfo{person}{Bo Chen},
  \bibinfo{person}{Ruoming Pang}, \bibinfo{person}{Vijay Vasudevan},
  \bibinfo{person}{Mark Sandler}, \bibinfo{person}{Andrew Howard}, {and}
  \bibinfo{person}{Quoc~V Le}.} \bibinfo{year}{2019}\natexlab{}.
\newblock \showarticletitle{{MNasNet}: Platform-aware neural architecture
  search for mobile}. In \bibinfo{booktitle}{\emph{Proceedings of the IEEE
  Conference on Computer Vision and Pattern Recognition (CVPR)}}.
  \bibinfo{pages}{2820--2828}.
\newblock


\bibitem[\protect\citeauthoryear{Tan and Le}{Tan and Le}{2019}]%
        {tan2019efficientnet}
\bibfield{author}{\bibinfo{person}{Mingxing Tan} {and} \bibinfo{person}{Quoc~V
  Le}.} \bibinfo{year}{2019}\natexlab{}.
\newblock \showarticletitle{{EfficientNet}: Rethinking model scaling for
  convolutional neural networks}. In \bibinfo{booktitle}{\emph{Proceedings of
  the International Conference on Machine Learning (ICML)}}.
\newblock


\bibitem[\protect\citeauthoryear{Tan, Pang, and Le}{Tan et~al\mbox{.}}{2020}]%
        {tan2019efficientdet}
\bibfield{author}{\bibinfo{person}{Mingxing Tan}, \bibinfo{person}{Ruoming
  Pang}, {and} \bibinfo{person}{Quoc~V Le}.} \bibinfo{year}{2020}\natexlab{}.
\newblock \showarticletitle{{EfficientDet}: Scalable and efficient object
  detection}.
\newblock  (\bibinfo{year}{2020}), \bibinfo{pages}{10781--10790}.
\newblock


\bibitem[\protect\citeauthoryear{Tancreti, Hossain, Bagchi, and
  Raghunathan}{Tancreti et~al\mbox{.}}{2011}]%
        {tancreti2011aveksha}
\bibfield{author}{\bibinfo{person}{Matthew Tancreti},
  \bibinfo{person}{Mohammad~Sajjad Hossain}, \bibinfo{person}{Saurabh Bagchi},
  {and} \bibinfo{person}{Vijay Raghunathan}.} \bibinfo{year}{2011}\natexlab{}.
\newblock \showarticletitle{Aveksha: A hardware-software approach for
  non-intrusive tracing and profiling of wireless embedded systems}. In
  \bibinfo{booktitle}{\emph{Proceedings of the 9th ACM Conference on Embedded
  Networked Sensor Systems}}. \bibinfo{pages}{288--301}.
\newblock


\bibitem[\protect\citeauthoryear{Teerapittayanon, McDanel, and
  Kung}{Teerapittayanon et~al\mbox{.}}{2016}]%
        {teerapittayanon2016branchynet}
\bibfield{author}{\bibinfo{person}{Surat Teerapittayanon},
  \bibinfo{person}{Bradley McDanel}, {and} \bibinfo{person}{HT Kung}.}
  \bibinfo{year}{2016}\natexlab{}.
\newblock \showarticletitle{{BranchyNet}: Fast inference via early exiting from
  deep neural networks}. In \bibinfo{booktitle}{\emph{Proceedings of IEEE
  International Conference on Pattern Recognition (ICPR)}}.
  \bibinfo{pages}{2464--2469}.
\newblock


\bibitem[\protect\citeauthoryear{Veit and Belongie}{Veit and Belongie}{2019}]%
        {veit2018convolutional}
\bibfield{author}{\bibinfo{person}{Andreas Veit} {and} \bibinfo{person}{Serge
  Belongie}.} \bibinfo{year}{2019}\natexlab{}.
\newblock \showarticletitle{Convolutional networks with adaptive inference
  graphs}.
\newblock \bibinfo{journal}{\emph{International Journal of Computer Vision}}
  \bibinfo{volume}{128} (\bibinfo{year}{2019}), \bibinfo{pages}{730--741}.
\newblock


\bibitem[\protect\citeauthoryear{Wang, Li, and Ling}{Wang
  et~al\mbox{.}}{2018}]%
        {wang2018pelee}
\bibfield{author}{\bibinfo{person}{Robert~J Wang}, \bibinfo{person}{Xiang Li},
  {and} \bibinfo{person}{Charles~X Ling}.} \bibinfo{year}{2018}\natexlab{}.
\newblock \showarticletitle{{PELEE}: A real-time object detection system on
  mobile devices}. In \bibinfo{booktitle}{\emph{Proceedings of the Advances in
  Neural Information Processing Systems (NeurIPS)}}.
  \bibinfo{pages}{1963--1972}.
\newblock


\bibitem[\protect\citeauthoryear{Wu, Dai, Zhang, Wang, Sun, Wu, Tian, Vajda,
  Jia, and Keutzer}{Wu et~al\mbox{.}}{2019}]%
        {wu2019fbnet}
\bibfield{author}{\bibinfo{person}{Bichen Wu}, \bibinfo{person}{Xiaoliang Dai},
  \bibinfo{person}{Peizhao Zhang}, \bibinfo{person}{Yanghan Wang},
  \bibinfo{person}{Fei Sun}, \bibinfo{person}{Yiming Wu},
  \bibinfo{person}{Yuandong Tian}, \bibinfo{person}{Peter Vajda},
  \bibinfo{person}{Yangqing Jia}, {and} \bibinfo{person}{Kurt Keutzer}.}
  \bibinfo{year}{2019}\natexlab{}.
\newblock \showarticletitle{{FBNet}: Hardware-aware efficient convnet design
  via differentiable neural architecture search}. In
  \bibinfo{booktitle}{\emph{Proceedings of the IEEE Conference on Computer
  Vision and Pattern Recognition (CVPR)}}. \bibinfo{pages}{10734--10742}.
\newblock


\bibitem[\protect\citeauthoryear{Wu, Nagarajan, Kumar, Rennie, Davis, Grauman,
  and Feris}{Wu et~al\mbox{.}}{2018}]%
        {wu2018blockdrop}
\bibfield{author}{\bibinfo{person}{Zuxuan Wu}, \bibinfo{person}{Tushar
  Nagarajan}, \bibinfo{person}{Abhishek Kumar}, \bibinfo{person}{Steven
  Rennie}, \bibinfo{person}{Larry~S Davis}, \bibinfo{person}{Kristen Grauman},
  {and} \bibinfo{person}{Rogerio Feris}.} \bibinfo{year}{2018}\natexlab{}.
\newblock \showarticletitle{{BlockDrop}: Dynamic inference paths in residual
  networks}. In \bibinfo{booktitle}{\emph{Proceedings of the IEEE Conference on
  Computer Vision and Pattern Recognition (CVPR)}}.
  \bibinfo{pages}{8817--8826}.
\newblock


\bibitem[\protect\citeauthoryear{Xu, Koo, Kumar, Bai, Mitra, Misailovic, and
  Bagchi}{Xu et~al\mbox{.}}{2018a}]%
        {xu2018videochef}
\bibfield{author}{\bibinfo{person}{Ran Xu}, \bibinfo{person}{Jinkyu Koo},
  \bibinfo{person}{Rakesh Kumar}, \bibinfo{person}{Peter Bai},
  \bibinfo{person}{Subrata Mitra}, \bibinfo{person}{Sasa Misailovic}, {and}
  \bibinfo{person}{Saurabh Bagchi}.} \bibinfo{year}{2018}\natexlab{a}.
\newblock \showarticletitle{{VideoChef}: efficient approximation for streaming
  video processing pipelines}. In \bibinfo{booktitle}{\emph{Proceedings of the
  USENIX Annual Technical Conference (USENIX ATC)}}. \bibinfo{pages}{43--56}.
\newblock


\bibitem[\protect\citeauthoryear{Xu, Mitra, Rahman, Bai, Zhou, Bronevetsky, and
  Bagchi}{Xu et~al\mbox{.}}{2018b}]%
        {xu2018pythia}
\bibfield{author}{\bibinfo{person}{Ran Xu}, \bibinfo{person}{Subrata Mitra},
  \bibinfo{person}{Jason Rahman}, \bibinfo{person}{Peter Bai},
  \bibinfo{person}{Bowen Zhou}, \bibinfo{person}{Greg Bronevetsky}, {and}
  \bibinfo{person}{Saurabh Bagchi}.} \bibinfo{year}{2018}\natexlab{b}.
\newblock \showarticletitle{Pythia: Improving datacenter utilization via
  precise contention prediction for multiple co-located workloads}. In
  \bibinfo{booktitle}{\emph{Proceedings of the International Middleware
  Conference (Middleware)}}. \bibinfo{pages}{146--160}.
\newblock


\bibitem[\protect\citeauthoryear{Yang, Han, Chen, Song, Dai, and Huang}{Yang
  et~al\mbox{.}}{2020}]%
        {yang2020resolution}
\bibfield{author}{\bibinfo{person}{Le Yang}, \bibinfo{person}{Yizeng Han},
  \bibinfo{person}{Xi Chen}, \bibinfo{person}{Shiji Song},
  \bibinfo{person}{Jifeng Dai}, {and} \bibinfo{person}{Gao Huang}.}
  \bibinfo{year}{2020}\natexlab{}.
\newblock \showarticletitle{Resolution Adaptive Networks for Efficient
  Inference}. In \bibinfo{booktitle}{\emph{Proceedings of the IEEE Conference
  on Computer Vision and Pattern Recognition (CVPR)}}.
  \bibinfo{pages}{2369--2378}.
\newblock


\bibitem[\protect\citeauthoryear{Zhang, Ananthanarayanan, Bodik, Philipose,
  Bahl, and Freedman}{Zhang et~al\mbox{.}}{2017}]%
        {videostorm}
\bibfield{author}{\bibinfo{person}{Haoyu Zhang}, \bibinfo{person}{Ganesh
  Ananthanarayanan}, \bibinfo{person}{Peter Bodik}, \bibinfo{person}{Matthai
  Philipose}, \bibinfo{person}{Paramvir Bahl}, {and} \bibinfo{person}{Michael~J
  Freedman}.} \bibinfo{year}{2017}\natexlab{}.
\newblock \showarticletitle{Live Video Analytics at Scale with Approximation
  and Delay-Tolerance.}. In \bibinfo{booktitle}{\emph{Proceedings of the
  Symposium on Networked Systems Design and Implementation (NSDI)}},
  Vol.~\bibinfo{volume}{9}. \bibinfo{pages}{377--392}.
\newblock


\bibitem[\protect\citeauthoryear{Zhang, Zhou, Lin, and Sun}{Zhang
  et~al\mbox{.}}{2018}]%
        {zhang2018shufflenet}
\bibfield{author}{\bibinfo{person}{Xiangyu Zhang}, \bibinfo{person}{Xinyu
  Zhou}, \bibinfo{person}{Mengxiao Lin}, {and} \bibinfo{person}{Jian Sun}.}
  \bibinfo{year}{2018}\natexlab{}.
\newblock \showarticletitle{{ShuffleNet}: An extremely efficient convolutional
  neural network for mobile devices}. In \bibinfo{booktitle}{\emph{Proceedings
  of the IEEE Conference on Computer Vision and Pattern Recognition (CVPR)}}.
  \bibinfo{pages}{6848--6856}.
\newblock


\bibitem[\protect\citeauthoryear{Zhao, Zheng, Xu, and Wu}{Zhao
  et~al\mbox{.}}{2019}]%
        {zhao2019object}
\bibfield{author}{\bibinfo{person}{Zhong-Qiu Zhao}, \bibinfo{person}{Peng
  Zheng}, \bibinfo{person}{Shou-tao Xu}, {and} \bibinfo{person}{Xindong Wu}.}
  \bibinfo{year}{2019}\natexlab{}.
\newblock \showarticletitle{Object detection with deep learning: A review}.
\newblock \bibinfo{journal}{\emph{IEEE Transactions on Neural Networks and
  Learning Systems}} \bibinfo{volume}{30}, \bibinfo{number}{11}
  (\bibinfo{year}{2019}), \bibinfo{pages}{3212--3232}.
\newblock


\bibitem[\protect\citeauthoryear{Zhu, Dai, Yuan, and Wei}{Zhu
  et~al\mbox{.}}{2018}]%
        {zhu2018towards}
\bibfield{author}{\bibinfo{person}{Xizhou Zhu}, \bibinfo{person}{Jifeng Dai},
  \bibinfo{person}{Lu Yuan}, {and} \bibinfo{person}{Yichen Wei}.}
  \bibinfo{year}{2018}\natexlab{}.
\newblock \showarticletitle{Towards high performance video object detection}.
  In \bibinfo{booktitle}{\emph{Proceedings of the IEEE Conference on Computer
  Vision and Pattern Recognition (CVPR)}}. \bibinfo{pages}{7210--7218}.
\newblock


\bibitem[\protect\citeauthoryear{Zhu, Wang, Dai, Yuan, and Wei}{Zhu
  et~al\mbox{.}}{2017a}]%
        {zhu2017flow}
\bibfield{author}{\bibinfo{person}{Xizhou Zhu}, \bibinfo{person}{Yujie Wang},
  \bibinfo{person}{Jifeng Dai}, \bibinfo{person}{Lu Yuan}, {and}
  \bibinfo{person}{Yichen Wei}.} \bibinfo{year}{2017}\natexlab{a}.
\newblock \showarticletitle{Flow-guided feature aggregation for video object
  detection}. In \bibinfo{booktitle}{\emph{Proceedings of the IEEE
  International Conference on Computer Vision (ICCV)}}.
  \bibinfo{pages}{408--417}.
\newblock


\bibitem[\protect\citeauthoryear{Zhu, Xiong, Dai, Yuan, and Wei}{Zhu
  et~al\mbox{.}}{2017b}]%
        {zhu2017deep}
\bibfield{author}{\bibinfo{person}{Xizhou Zhu}, \bibinfo{person}{Yuwen Xiong},
  \bibinfo{person}{Jifeng Dai}, \bibinfo{person}{Lu Yuan}, {and}
  \bibinfo{person}{Yichen Wei}.} \bibinfo{year}{2017}\natexlab{b}.
\newblock \showarticletitle{Deep feature flow for video recognition}. In
  \bibinfo{booktitle}{\emph{Proceedings of the IEEE Conference on Computer
  Vision and Pattern Recognition (CVPR)}}. \bibinfo{pages}{2349--2358}.
\newblock


\bibitem[\protect\citeauthoryear{Zilberstein}{Zilberstein}{1996}]%
        {zilberstein1996using}
\bibfield{author}{\bibinfo{person}{Shlomo Zilberstein}.}
  \bibinfo{year}{1996}\natexlab{}.
\newblock \showarticletitle{Using anytime algorithms in intelligent systems}.
\newblock \bibinfo{journal}{\emph{AI magazine}} \bibinfo{volume}{17},
  \bibinfo{number}{3} (\bibinfo{year}{1996}), \bibinfo{pages}{73--73}.
\newblock


\end{thebibliography}

\end{document}